\pdfoutput=1

\documentclass[review]{elsarticle}

\usepackage{amsmath}
\usepackage{amssymb}
\usepackage{commath}
\usepackage{comment}
\usepackage{graphicx}
\usepackage{booktabs}
\usepackage{multirow}
\usepackage{colortbl}
\usepackage{tabularx}

\usepackage{subfig}
\usepackage{lineno}
\usepackage{hyperref}
\usepackage{mathrsfs}
\usepackage[T1]{fontenc}
\usepackage{frcursive}
\usepackage{calligra}

\usepackage{bm}

\usepackage{algorithm}
\usepackage{algpseudocode}
\makeatletter
\def\BState{\State\hskip-\ALG@thistlm}
\makeatother

\modulolinenumbers[5]

\biboptions{authoryear}

\begin{document}

\begin{frontmatter}

\title{Short-Term Forecasting of Passenger Demand under On-Demand Ride Services: A Spatio-Temporal Deep Learning Approach}

%%% Group authors per affiliation:

\author[mymainaddress]{Jintao~Ke}
\author[mysecondaryaddress]{Hongyu Zheng}
\author[mymainaddress]{Hai Yang}

\author[mysecondaryaddress]{Xiqun~(Michael)~Chen\corref{mycorrespondingauthor}}
\cortext[mycorrespondingauthor]{Corresponding author}
\ead{chenxiqun@zju.edu.cn}

\address[mymainaddress]{Department of Civil and Environmental Engineering, Hong Kong University of Science and Technology, Clear Water Bay, Kowloon, Hong Kong, China}
\address[mysecondaryaddress]{College of Civil Engineering and Architecture, Zhejiang University, Hangzhou, China}

\begin{abstract}
Short-term passenger demand forecasting is of great importance to the on-demand ride service platform, which can incentivize vacant cars moving from over-supply regions to over-demand regions. The spatial dependences, temporal dependences, and exogenous dependences need to be considered simultaneously, however, which makes short-term passenger demand forecasting challenging. We propose a novel deep learning (DL) approach, named the fusion convolutional long short-term memory network (FCL-Net), to address these three dependences within one end-to-end learning architecture. The model is stacked and fused by multiple convolutional long short-term memory (LSTM) layers, standard LSTM layers, and convolutional layers. The fusion of convolutional techniques and the LSTM network enables the proposed DL approach to better capture the spatio-temporal characteristics and correlations of explanatory variables. A tailored spatially aggregated random forest is employed to rank the importance of the explanatory variables. The ranking is then used for feature selection. The proposed DL approach is applied to the short-term forecasting of passenger demand under an on-demand ride service platform in Hangzhou, China. Experimental results, validated on real-world data provided by DiDi Chuxing, show that the FCL-Net achieves better predictive performance than traditional approaches including both classical time-series prediction models and neural network based algorithms (e.g., artificial neural network and LSTM). Furthermore, the consideration of exogenous variables in addition to passenger demand itself, such as the travel time rate, time-of-day, day-of-week, and weather conditions, is proven to be promising, since it reduces the root mean squared error (RMSE) by 50.9\%. It is also interesting to find that the feature selection reduces 30\% in the dimension of predictors and leads to only 0.6\% loss in the forecasting accuracy measured by RMSE in the proposed model. This paper is one of the first DL studies to forecast the short-term passenger demand of an on-demand ride service platform by examining the spatio-temporal correlations.
\end{abstract}


\end{frontmatter}

% \linenumbers

\section{Introduction}
The on-demand ride service platform, e.g., Urber, Lyft, DiDi Chuxing, is an emerging technology with the boom of the mobile internet. Ride-sourcing or transportation network companies (TNCs) refer to an emerging urban mobility service mode that private car owners drive their own vehicles to provide for-hire rides\citep{chen2017understanding}. On-demand ride-sourcing services can be completed via smartphone applications. The platform serves as a coordinator who matches requesting orders from passengers (demand) and vacant registered cars (supply). There exists an abundance of leverages to influence drivers' and passengers' preference and behavior, and thus affect both the demand and supply, to maximize profits of the platform or achieve maximum social welfare. Having better understanding of the short-term passenger demand over different spatial zones is of great importance to the platform or the operator, who can incentivize drivers to the zones with more potential passenger demands, and improve the utilization rate of the registered cars.

Although limited research efforts have been implemented on forecasting short-term passenger demand under the emerging on-demand ride service platform in most recent years mainly due to the real-world data unavailability, the fruitful studies on the taxi market can provide valuable insights since there exist strong similarities between the taxi market and the on-demand ride service market. A series of mathematic models were developed to spell out endogenous relationships among variables in the taxi market \citep{yang2010equilibria,yang2011equilibrium,yang2002demand,yang2005multiperiod} under the two-sided market equilibrium. On the demand side, the accurate passenger demand was affected by passengers' waiting time and taxi fare; while on the supply side, drivers' behavior, i.e., how to find a passenger, was mainly affected by the expected searching time and taxi fare. The passenger demand was endogenously determined when the taxi operator decided the taxi fare structure and the number of released licenses of taxis (entry limitation).

In theory, the equilibrium between the demand and supply will eventually be reached when the arrival rate of passengers equals to the arrival rate of vacant taxis and equals to the meeting rate. However, heterogeneous and exogenous factors in reality, e.g., asymmetric information, short-term fluctuations, may make it difficult to guarantee the spatial distribution of taxis matching the passenger demand all the time \citep{moreira2013predicting}. Hence, disequilibrium states can result from the following two scenarios: oversupply (an excess in the number of vacant taxis may decrease the taxi utilization) and overfull demand (excessively waiting passengers may lower the degree of satisfaction). Both scenarios are harmful to the taxi operator as well as the on-demand ride service platform, raising a strong need for a precise forecasting of short-term passenger demand. It helps the operator/platform implement proactive incentive mechanism, such as surge pricing and cash/point awards, to attract drivers from regions of oversupply to regions with overfull demand. These strategies not only shorten the process of reaching equilibrium under a dynamic environment but also help improve the taxi/car utilization rate and reduce passengers' waiting time.

However, short-term forecasting of passenger demand or on-demand ride services in each region is of great challenge mainly due to three kinds of dependences \citep{zhang2016deep}:

\begin{enumerate}[(1)]
  \item Time dependences: passenger demand has a strong periodicity (for example, the passenger demand is expected to be high during morning and evening peaks and to be low during sleeping hours); furthermore, the short-term passenger demand is dependent on the trend of the nearest historical passenger demand.
  \item Spatial dependences: \cite{yang2010equilibria} revealed that the passenger demand in one specific zone was not merely determined by the variables of this zone, but endogenously dependent on all the zonal variables in the whole network. Generally, the variables of the nearby zones have stronger influences than distant zones, which inspires the need for an advanced model that can capture local spatial dependencies.
  \item Exogenous dependences: some exogenous variables, such as the travel time rate and weather conditions, may have strong influences on the short-term passenger demand. The exogenous variables also demonstrate time dependences and spatial dependences.
\end{enumerate}

Although little direct experience suggests solutions to these three dependences in short-term passenger demand forecasting, studies on traffic speed/volume prediction and rainfall nowcasting provide valuable insights \citep{ghosh2009multivariate, huang2009novel,guo2014adaptive,wang2014new}. Recently, deep learning (DL) approaches have been successfully used for traffic flow prediction. For example, \cite{ma2015long} employed the long short-term memory (LSTM) neural network to capture the long-term dependences and nonlinear traffic dynamics for short-term traffic speed prediction. \cite{wu2016short} incorporated 1-dimension convolutional neural network (CNN) and LSTM in short-term traffic flow forecasting in order to capture spatio-temporal correlations. \cite{zhang2016deep} presented a deep spatio-temporal residual network to predict the inflow and outflow in each region of a city simultaneously. \cite{xingjian2015convolutional} innovatively integrated CNN and LSTM in one end-to-end DL structure, named the convolutional LSTM (conv-LSTM), which provided a brand-new idea for solving spatio-temporal sequence forecasting problems. In that research, numerical experiments showed that the conv-LSTM outperformed fully connected LSTM in two datasets.

In this paper, we propose a novel DL structure, named the fusion convolutional LSTM network (FCL-Net), to consider the three dependences simultaneously in the short-term passenger demand forecasting for the on-demand ride service platform. Different from aforementioned studies, this structure coordinates the spatio-temporal variables and non-spatial time-series variables in one end-to-end trainable model. Before feeding these explanatory variables into the DL structure, a tailored spatial aggregated random forest is designed to evaluate the feature importance with different categories, look-back time intervals, and spatial locations.

To the best knowledge of the authors, this paper is one of the first attempts to employ spatio-temporal DL approaches in short-term passenger demand forecasting under the on-demand ride service platform. The main contributions of this paper are within three folds:

\begin{enumerate}[(1)]
  \item The novel FCL-Net approach characterizes the spatio-temporal properties of the predictors, captures the temporal features of non-spatial time-series variables simultaneously, and coordinates them in one end-to-end learning structure for the short-term passenger demand forecasting.
  \item We extract the potential predictors affecting short-term passenger demand and assess the feature importance of these predictors via a spatial aggregated random forest.
  \item Validated by the real-world on-demand ride services data provided by DiDi Chuxing in a large-scale urban network, the proposed DL structure outperforms five benchmark algorithms, including three conventional time-series prediction methods and two classical DL algorithms.
\end{enumerate}

The rest of the paper is organized as follows. Section 2 first reviews the existing research on the taxi market modeling and taxi-passenger demand forecasting, and then summarizes the state-of-the-art DL approaches utilized in related problems. Section 3 formulates the short-term traffic flow forecasting problem, and explicitly explains the explanatory variables. Section 4 describes the structure and mathematical formulation of the proposed FCL-Net, as well as the proposed spatial aggregated random forest algorithm for feature selection. Section 5 compares the predictive performance between the proposed approach and the benchmark models, including the historical average (HA), moving average (MA), autoregressive integrated moving average (ARIMA), and two classical DL algorithms (artificial neural network and LSTM), based on the real-world dataset extracted from DiDi Chuxing. Finally, Section 6 concludes the paper and outlooks the future research.

\section{Literature Review}

The fast-growing technology of mobile internet enables on-demand ride service platforms for providing efficient connections between waiting passengers and vacant registered cars \citep{tang2016coordinating}. Like the taxi market, the on-demand ride service market is essentially a two-sided market where both the consumers (passengers) and providers (drivers of vacant cars) are independent and have individual mode choices. The passengers make mode choice decisions between taxi/on-demand ride service and public transportation according to the waiting time and trip fare, while the drivers make service decisions by considering the searching time and trip fare. In light of the fact that the vacant taxis and waiting passengers are unable to be matched simultaneously in a specific zone, \citet{yang2002demand,yang2010equilibria,yang2011equilibrium} proposed a meeting function to characterize the search frictions between drivers of vacant taxis and waiting passengers. The meeting function pointed out that the meeting rate in one specific zone was determined by the density of the waiting passengers and vacant taxis at that moment, which indicated that passengers' waiting time, drivers' searching time, and passengers' arrival rate (demand) were endogenously correlated. The equilibrium state was reached when the arrival rate of waiting passengers exactly matched the arrival rate of vacant taxis. This equilibrium state along with the endogenous variables were influenced by the exogenous variables, such as the taxi fleet size and taxi trip fare. The taxi operator might coordinate supply and demand via the on-demand service platform and thus influence the equilibrium state by regulating the entry of taxi and determining the taxi fare structure, such as non-linear pricing \citep{yang2010nonlinear}. Apart from the traditional taxi market, some emerging market structures, like the ride-sourcing market \citep{zha2016economic}, e-hailing taxi market \citep{he2015modeling,wang2016pricing}, were examined under the same equilibrium modeling framework. Recently, the on-demand ride-sharing market and the optimal assignment strategies have also attracted researchers' attentions\citep{alonso2017demand}. 

However, researchers found that a regional disequilibrium occurred when there was an excess in vacant taxis or waiting passengers in that region \citep{moreira2012predictive}. This disequilibrium might lead to a resource mismatch between supply and demand, which resulted in low taxi utilization in some regions while low taxi availability in other regions. Therefore, a short-term passenger demand forecasting model is of great importance to the taxi operator, which can implement efficient taxi dispatching and time-saving route finding to achieve an equilibrium across urban regions \citep{zhang2017taxi}. To attain the accurate and robust short-term passenger demand forecasting, both parametric (e.g., ARIMA) and non-parametric models (e.g., neural network) have been examined. For instance, \cite{zhao2016predicting} implemented and compared three models, i.e., the Markov algorithm, Lempel-Ziv-Welch algorithm, and neural network. In that research, the results showed that neural network performed better with the lower theoretical maximum predictability while the Markov predictor had better performance with the higher theoretical maximum predictability. \cite{moreira2013predicting} proposed a data stream ensemble framework which incorporated time varying passion model and ARIMA, to predict the spatial distribution of taxi passenger demand. \cite{deng2011spatiotemporal} employed the global and local Moran's I values to evaluate the intensity of taxi services in Shanghai. Some socio-demographical and built-environment variables have also been in use for predicting taxi passenger demand \citep{qian2015spatial}. 

There are a broad range of problems in the domain of transportation, which are similar to short-term passenger demand forecasting. These problems include the traffic speed estimation \citep{bachmann2013comparative,soriguera2011estimation,wang2013short}, traffic volume prediction \citep{boto2010wavelet}, real-time crash likelihood estimation \citep{ahmed2013data,yu2014utilizing}, human mobility pattern forecasting \citep{ouyang2016deepspace}, car-following behavior prediction\citep{Wang2017cap}, original-destination matrices forecasting \citep{toque2016forecasting}, bus arrival time prediction \citep{yu2011bus}, short-term forecasting of high speed rail demand \citep{jiang2014short}, and etc., the solutions to which offer meritorious inspirations to our problem. To solve these spatio-temporal forecasting problems, a broad range of approaches have been proposed, including the ARIMA family \citep{zhang2011seasonal,khashei2012hybridization}, local regression model \citep{antoniou2013dynamic}, neural network based algorithms \citep{chan2012neural}, and Bayesian inferring approaches \citep{fei2011bayesian}. \cite{vlahogianni2014short} reviewed the existing literature on short-term traffic forecasting, and observed that researchers were moving from classical statistic models to neural network based approaches with the explosive growth of data accessibility and computing power.

Recently, more and more DL algorithms have been utilized in traffic prediction due to their capability of capturing complex relationship from a huge amount of data. \cite{chenganalysis} proposed a DL based approach to forecast day-to-day travel demand variations in a large-scale traffic network. \cite{huang2014deep} predicted short-term traffic flow via a two-layer DL structure with a deep belief network (DBN) at the bottom and a multitask regression model (MTL) at the top. \cite{polson2017deep} found that the sharp nonlinearities of traffic flow, as a result of transitions between the free flow, breakdown, recovery and congestion, could be captured by a DL architecture. Combining the empirical mode decomposition (EMD) and back-propagation neural network (BPN), \cite{wei2012forecasting} presented a hybrid EMD-BPN method for short-term passenger flow forecasting. Graphical LASSO was also combined in the neural network, showing its potential in network-scale traffic flow forecasting \citep{sun2012network}. \cite{lv2015traffic} stated that a stacked autoencoder model helped to capture generic traffic flow features and characterize spatial temporal correlations in traffic flow prediction.

One of the obstacles in traffic forecasting is how to capture spatio-temporal correlations. It was found that the vehicle accumulation and dissipation had impacts on the travel volume of adjacent links or intersections, which indicated the spatial correlations should be considered in forecasting \citep{zhu2014traffic}. In terms of the spatial correlations, CNN developed by \citep{lecun1999object} was used to learn the local and global spatial correlations in large-scale, network-wide traffic forecasting \citep{chen2016learning}. To address temporal correlations (another inherit property in real-time traffic forecasting), the family of recurrent neural networks (RNN) \citep{williams1989learning} was widely viewed as one of the most suitable structures \citep{zhao2017lstm}. In the RNN architecture, the dependent variable in one timestamp was not only dependent on the explanatory variables in this timestamp, but also correlated with the explanatory variables in the previous timestamps \citep{graves2013generating,sutskever2009recurrent}. However, the traditional RNN suffered from a ``vanishing gradience'' effect which made it impossible to store long-term information \citep{hochreiter1991untersuchungen}. To address this issue, \cite{hochreiter1997long} presented the long short-term memory (LSTM) which employed a series of memory cells to store information for exploring long-range dependences in the data.

However, neither CNN nor LSTM are perfect models for spatio-temporal forecasting problems. CNN fails to capture the temporal dependences while LSTM is incapable of characterizing local spatial correlations. To capture spatial and temporal dependences simultaneously in one end-to-end training model, researchers have made numerous attempts in recent years. \cite{wang2016traffic} proposed a novel error-feedback recurrent convolutional neural network (eRCNN) architecture which was comprised of the input layer, the convolutional layer, and the error-feedback recurrent layer. \cite{zhang2016deep} modeled the temporal closeness, period, and trend properties of the inflow/outflow of human mobilities with serval separate convolutional layers and then fused these layers in one end-to-end DL structure. \cite{xingjian2015convolutional} proposed the conv-LSTM network, which combined CNN and LSTM in one sequence to sequence learning framework, for precipitation nowcasting that was a typical spatio-temporal forecasting problem. In that research, the results showed that the conv-LSTM outperformed fully-connected LSTM, since some complicated spatio-temporal characteristics could be learnt by the convolution and recurrent structure of the model.

In this paper, considering that short-term passenger demand is not only dependent on its own spatio-temporal properties but also dependent on other explanatory variables (some with spatio-temporal properties and some only with temporal properties), we extend the structure of the conv-LSTM to a more generalized architecture which addresses spatial, temporal, and exogenous dependences at the same time.

\section{Preliminaries}

The short-term passenger demand forecasting is essentially a time-series prediction problem, which implies that the nearest historical passenger demand can be valuable information for predicting the future demand. We also observe that the travel time rate also influences the short-term passenger demand, since it reflects the congestion level of trips and zones. For example, passengers will potentially transfer to subways if they find the trips to their destinations are congested. Furthermore, the attributes of time-of-day, day-of-week, and weather conditions also have impacts on short-term passenger demand.

In this section, we first interpret the notations of the variables used in this paper, and then give an explicit definition of the short-term passenger demand forecasting problem.

\textbf{Definition 1} \emph{(Region and time partition)}: The urban area is partitioned into $I \times J$ grids uniformly where each grid refers to a zone. On the other hand, we consider variables aggregated in a one-hour time interval in this paper.

Based on Definition 1, we explicitly define several categories of variables as follows:

(1)	Demand intensity

The intensity of demand at the $t$th time slot (e.g., hour) lying in grid $(i,j)$ is defined as the number of orders during this time interval within the grid, which is denoted by $d_t^{i,j}$. The intensity of demand in all $I \times J$ grids at the $t$th time slot is defined as the matrix $\bm{D}_t\in{R^{I \times J}}$ ($R$ refers to the real set), where the  $(i,j)$th element is $\left(\bm{D}_t\right)_{i,j}$ = $d_t^{i,j}$.

(2)	Average travel time rate

The travel time rate represents the travel time per unit travel distance \citep{chen2017understanding}. In this paper, the travel time rate of the  $m$th order originating from grid $(i,j)$ at the  $t$th time slot, is defined as the ratio of its travel time to its travel distance, $\tau
_{t,m}^{i,j}$. The average travel time rate in grid $(i,j)$ during the $t$th time slot, $\tau_t^{i,j}$, is defined as the average of $\tau_{t,m}^{i,j}$ over $m$. The average travel time rate in all $I \times J$ grids at the  $t$th time slot is defined as the matrix $\mathbf{\Gamma}_t\in{R^{I \times J}}$, where the $(i,j)$th element is $\left(\mathbf{\Gamma}_t\right)_{i,j}$ = $\tau_t^{i,j}$.

(3)	Time-of-day and day-of-week

By empirically examining the distribution of demand intensity with respect to time in the training dataset, 24 hours in each day can be intuitively divided into 3 periods: peak hours, off-peak hours, and sleep hours. We simply rank the hours based on the empirical demand intensity, and define the top 8 hours, middle 8 hours, bottom 8 hours, as the peak hours, off-peak hours and sleep hours. We further introduce the dummy variable $h_t$ to characterize this attribute of time-of-day, given by

$$ h_t=\left\{
\begin{aligned}
2,  &      &\textrm{if} \ \ t \ \ \textrm{belongs to peak hours} \\
1,  &      &\textrm{if} \ \ t \ \ \textrm{belongs to off-peak hours} \\
0,  &      &\textrm{if} \ \ t \ \ \textrm{belongs to sleep hours}
\end{aligned}
\right.
$$

We also denote another dummy variable $w_t$ to be the day-of-week, which catches up the distinguished properties between weekdays and weekends.

$$ w_t=\left\{
\begin{aligned}
0,  &      &\textrm{if} \ \ t \ \ \textrm{belongs to weekdays} \\
1,  &      &\textrm{if} \ \ t \ \ \textrm{belongs to weekends} \\
\end{aligned}
\right.
$$

(4)	Weather

We consider 5 categories of weather variables, including temperature, humidity, weather state, wind speed, and visibility. Further, the weather state consists of 5 categories, including sunny (5), cloudy (4), light rain (3), moderate rain (2), and heavy rain (1). In this paper, the temperature, humidity, weather state, wind speed, and visibility during the $t$th time interval are denoted as $at_t,ah_t,as_t,aw_t,av_t$, respectively.

All of the aforementioned variables demonstrate time-varying attributes, but the demand intensity and average travel time rate show zonal-based attributes, which mean they have different values across grids. The variables with time-varying attributes only have temporal dependences, while the variables with time-varying and zonal-based attributes simultaneously have both spatial and temporal dependences, which implies that they should be treated in different ways. Thus, we give the definition of the spatio-temporal variables and non-spatial time-series variables in Definition 2.

\textbf{Definition 2} \emph{(Spatio-temporal variables)}: refer to the variables showing distinction across time and across space, which imply there exist spatio-temporal correlations, e.g., the demand intensity and travel time rate. Other variables, including the time-of-day, day-of-week, and weather variables, are denoted as non-spatial time-series variables, which vary across time instead of space.

With the aforementioned definition of the explanatory variables, we can formulate the short-term passenger demand forecasting as Problem 1.

\textbf{Problem 1}: Given the historical observations and pre-known information $\{\bm{D}_s,\bm{\Gamma}_s | s=0,..., |s=0,§Ò§ß,t-1;h_s,w_s |s=0,§Ò§ß,t;at_s,ah_s,as_s,aw_s,av_s |s=0,...,t-1\}$, predict $\bm{D}_t$. It is noteworthy that the time-of-day and day-of-week of the $t$th time slot ($h_t$ and $w_t$) are pre-known at $t$.

\section{Methodology}
In this paper, we propose a novel DL architecture, i.e., FCL-Net, to capture the spatial dependences, temporal dependences, and exogenous dependences, in short-term passenger demand forecasting. To reduce the computation complexity, we also present a spatial aggregated random forest algorithm to rank the importance of explanatory variables and select the important ones. In this section, we first present a brief review of the traditional LSTM and conv-LSTM, then introduce the proposed architecture and training algorithm of FCL-Net, and finally illustrates the proposed spatial aggregated random forest.

\subsection{LSTM and Conv-LSTM}

The traditional artificial neural network (ANN) lacks the ability to catch up time-series characteristics since it does not take the temporal dependences into consideration. To overcome this shortcoming, the RNN is proposed, where the connection between units is organized by timestamps. The inner structure of an RNN layer is illustrated in Figure 1, where the input is a $T$ time-stamp vector sequence $\bm{x}=(\bm{x}_1,\bm{x}_2,...,\bm{x}_T)$  and the output is a hidden vector sequence $\bm{h}=(\bm{h}_1,\bm{h}_2,...,\bm{h}_T)$. It is noteworthy that $\bm{x}_t$ can be a one-dimensional vector or scalar, while $\bm{h}_t$ does not necessarily have the same dimension as $\bm{x}_t$. The hidden unit value in timestamp $t$, i.e., $\bm{h}_{t}$, stores the information, including hidden values $(\bm{h}_1,\bm{h}_2,...,\bm{h}_{t-1} )$ and input values $(\bm{x}_1,\bm{x}_2,...,\bm{x}_{t-1} )$, of the previous timestamps. Together with the input in $t$, i.e., $\bm{x}_t$, it is passed to the next timestamp $t+1$ at each iteration. In this way, RNN can memorize the information from multiple previous timestamps. Although RNN exhibits strong ability in catching temporal characteristics, it fails to store information for a long-term memory.

%% figure 1
\begin{figure}[!t]
	\centering
	\includegraphics[width=0.8\linewidth,trim=4 4 4 4,clip]{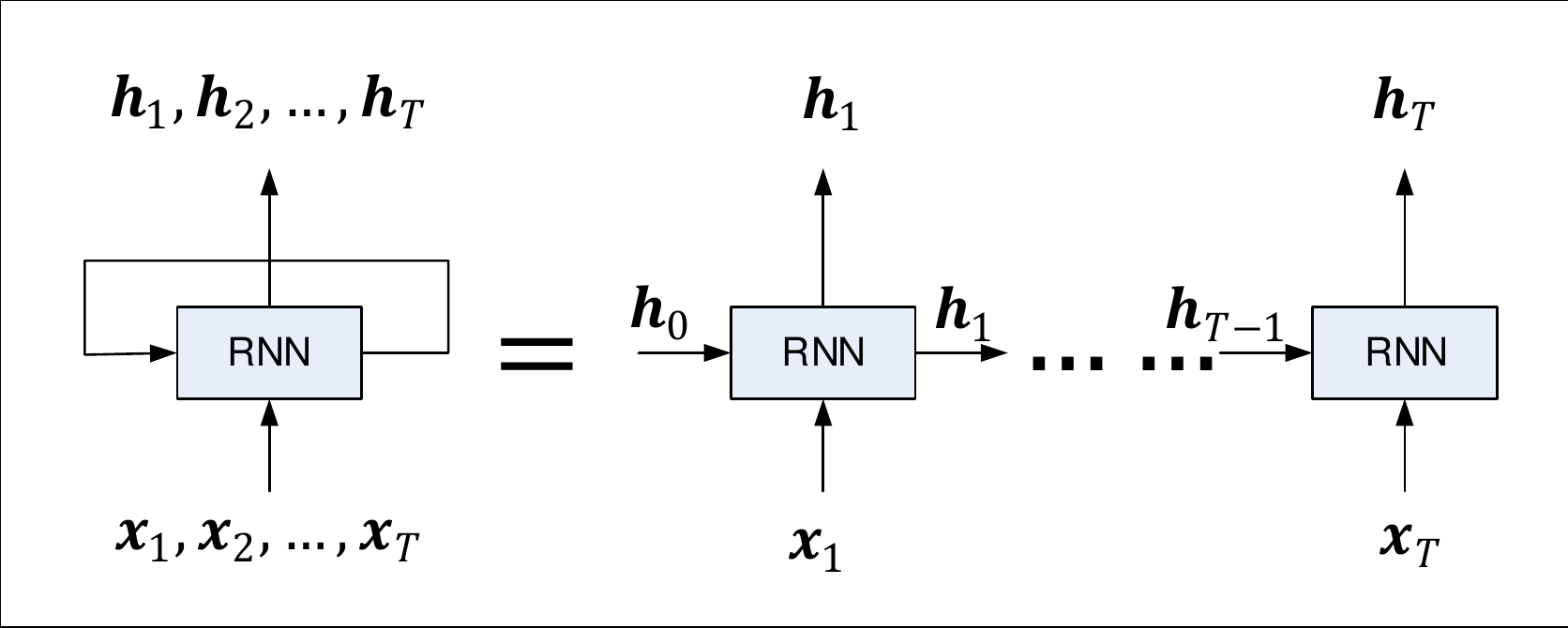}
	\caption{Illustration of the inner structure of an RNN layer.}
	\label{figure1}
\end{figure}

LSTM, as a special RNN structure, overcomes RNN's weakness on the long-term memory. Like the standard RNN, each LSTM cell maps the input vector sequence $\bm{x}$ to a hidden vector sequence $\bm{h}$ by $T$ iterations. As demonstrated in Eqs. (1)-(5), $\bm{i}_t,\bm{f}_t,\bm{o}_t,\bm{c}_t,( t=1,2,...,T)$ represent the input gate, forget gate, output gate, and memory cell vectors, respectively, sharing the same dimension with $\bm{h}_t$.

\begin{equation}
\bm{i}_t = \sigma(\bm{W}_{xi} \bm{x}_t + \bm{W}_{hi} \bm{h}_{t-1} + \bm{W}_{ci}\circ{\bm{c}_{t-1}} +\bm{b}_{i})
\end{equation}

\begin{equation}
\bm{f}_t = \sigma(\bm{W}_{xf} \bm{x}_t + \bm{W}_{hf} \bm{h}_{t-1} + \bm{W}_{cf}\circ{\bm{c}_{t-1}} +\bm{b}_{f})
\end{equation}

\begin{equation}
\bm{c}_t = \bm{f}_t\circ{\bm{c}_{t-1}} + \bm{i}_t\circ{\tanh{(\bm{W}_{xc} \bm{x}_t + \bm{W}_{hc} \bm{h}_{t-1} + \bm{b}_{c})}}
\end{equation}

\begin{equation}
\bm{o}_t = \sigma(\bm{W}_{xo} \bm{x}_t + \bm{W}_{ho} \bm{h}_{t-1} + \bm{W}_{co}\circ{\bm{c}_{t}} +\bm{b}_{o})
\end{equation}

\begin{equation}
\bm{h}_t = \bm{o}_t\circ{\tanh{\bm{c}_t}}
\end{equation}

The operator `$\circ$' refers to Hadamard product, which calculates the element-wise products of two vectors, matrices, or tensors with the same dimensions. $\sigma$ and $\tanh$ are the two non-linear activation functions given by

\begin{equation}
\sigma(x) = \dfrac{1}{1+e^{-x}}
\end{equation}

\begin{equation}
\tanh{x} = \dfrac{e^{x}-e^{-x}}{e^{x}+e^{-x}}
\end{equation}

$\bm{W}_{cf},\bm{W}_{ci},\bm{W}_{co},\bm{W}_{xi},\bm{W}_{hi},\bm{W}_{xf},\bm{W}_{hf},\bm{W}_{xc},\bm{W}_{hc},\bm{W}_{xo},\bm{W}_{ho}$ are the weighted parameter matrices which conduct a linear transformation from the vector of the first subscript to the second subscript, while $\bm{b}_i,\bm{b}_f,\bm{b}_c,\bm{b}_o$ are the intercept parameters.

Multiple LSTM cells can be stacked to form a deeper and more complicated neural network, which can better discover the complex relationships between the inputs and outputs.  In this paper, each LSTM cell is denoted as a function $\mathcal{F}^{L}:R^{T \times L} \to R^{T \times L'}$, where $T$ is the length of time sequences, $L$ is the length of one input vector, and $L'$ is the length of one output vector.

%% figure 2
\begin{figure}[!t]
	\centering
	\includegraphics[width=0.9\linewidth,trim=4 4 4 4,clip]{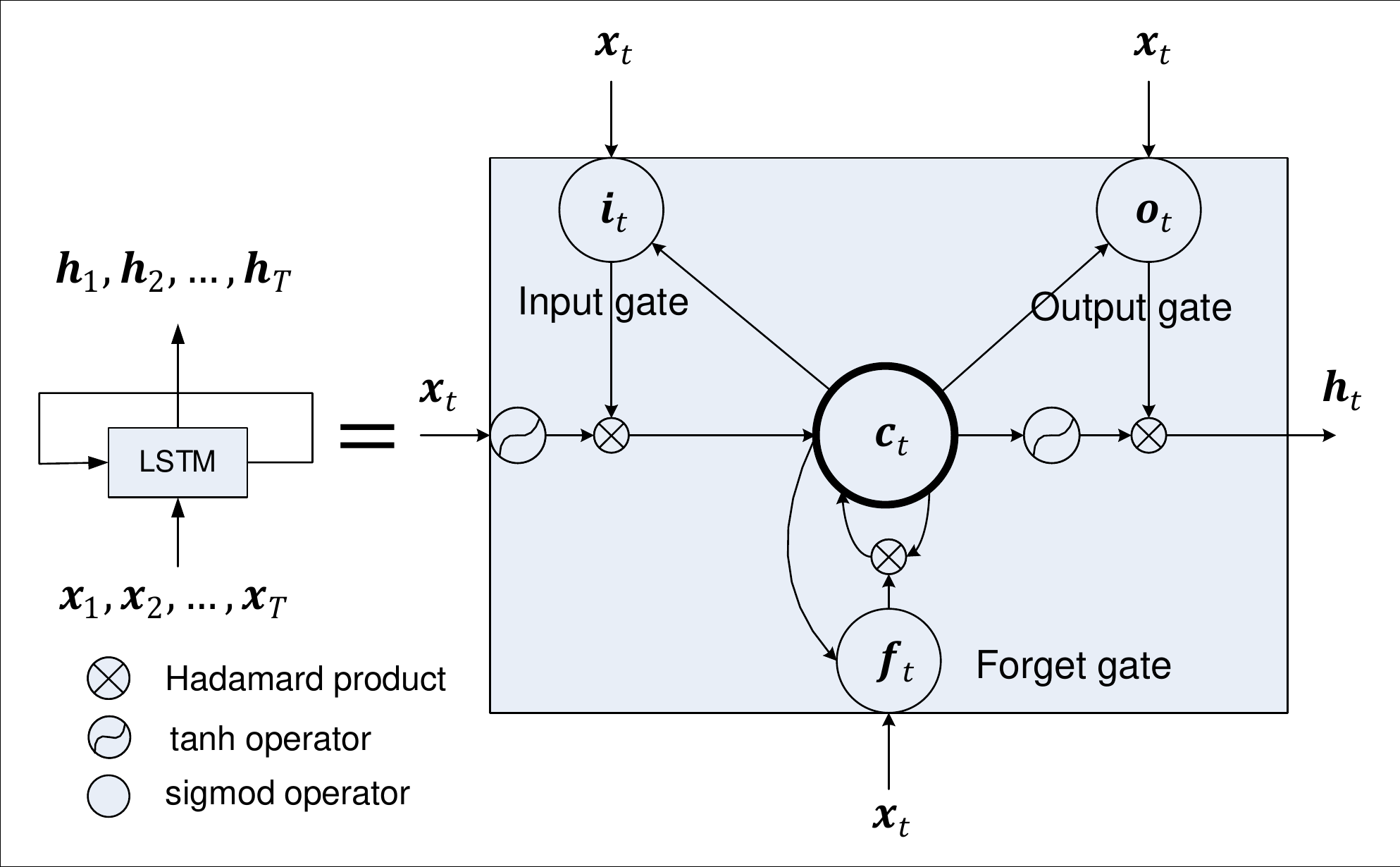}
	\caption{Illustration of the inner structure of an LSTM layer.}
	\label{figure2}
\end{figure}

However, LSTM is not an ideal model for the passenger demand forecasting with spatial and temporal characteristics in this paper, because it fails to capture the spatial dependences. To overcome this shortcoming, conv-LSTM network, which combines CNN and LSTM in one end-to-end DL architecture, is proposed.

The core idea of the conv-LSTM is to transform all the inputs, memory cell values, hidden states, and various gates in Eqs. (1)-(5) to 3D tensors (shown in Eqs. (8)-(12)).

\begin{equation}
\mathcal{I}_t = \sigma(\bm{W}_{xi}* \mathcal{X}_t + \bm{W}_{hi}* \mathcal{H}_{t-1} + \bm{W}_{ci}\circ{\mathcal{C}_{t-1}} +\bm{b}_{i})
\end{equation}

\begin{equation}
\mathcal{F}_t = \sigma(\bm{W}_{xf}* \mathcal{X}_t + \bm{W}_{hf}* \mathcal{H}_{t-1} + \bm{W}_{cf}\circ{\mathcal{C}_{t-1}} +\bm{b}_{f})
\end{equation}

\begin{equation}
\mathcal{C}_t = \mathcal{F}_t\circ{\mathcal{C}_{t-1}} + \mathcal{I}_t\circ{\tanh{(\bm{W}_{xc}* \mathcal{X}_t + \bm{W}_{hc}* \mathcal{H}_{t-1} + \bm{b}_{c})}}
\end{equation}

\begin{equation}
\mathcal{O}_t = \sigma(\bm{W}_{xo}* \mathcal{X}_t + \bm{W}_{ho}* \mathcal{H}_{t-1} + \bm{W}_{co}\circ{\mathcal{C}_{t}} +\bm{b}_{o})
\end{equation}

\begin{equation}
\mathcal{H}_t = \mathcal{O}_t\circ{\tanh{\mathcal{C}_t}}
\end{equation}

The input tensors, hidden tensors, memory cell tensors, input gate tensors, output gate tensors, and forget gate tensors are denoted as $\mathcal{X}_t,\mathcal{H}_t,\mathcal{C}_t,\mathcal{I}_t,\mathcal{O}_t,\mathcal{F}_t  \in R^{M \times N \times L}$, respectively, where $M,N$ are spatial dimensions ($M$ rows and $N$ columns of the grids). The operator $*$ stands for the convolutional operator. Here, $\bm{W}_{xf},\bm{W}_{hf},\bm{W}_{xc},\bm{W}_{hc},\bm{W}_{xo},\bm{W}_{ho}$ serve as convolutional flitters, which are replicated across the tensors with shared weights, and thus explore spatially local correlations. To maintain the consistence of the spatial dimensions (rows and columns), zero padding is employed before applying the convolutional operator.

Through these $T$ iterations, each conv-LSTM layer can map a sequence of input tensors $\mathcal{X}=(\mathcal{X}_1,\mathcal{X}_2,...,\mathcal{X}_T)$ to a sequence of hidden tensors $\mathcal{H}=(\mathcal{H}_1,\mathcal{H}_2,...,\mathcal{H}_T)$. In this paper, each conv-LSTM cell is denoted as a function $\mathcal{F}^{CL}:R^{T \times M \times N \times L} \to R^{T \times M \times N \times L'}$, where $T$ is the length of time sequences, $M,N$ refer to dimensions of rows and columns, respectively. Similar to LSTM, multiple conv-LSTM layers can be stacked to build up a deep conv-LSTM neural network.

However, the spatio-temporal variables, such as demand intensity and travel time rate during one time interval, are 2D matrices (see definition 1 and 2), thus a transformation function $\mathcal{F}^{T}:R^{M \times N} \to R^{M \times N \times 1}$ is employed to transfer the initial input matrices into 3D tensors by simply adding one dimension.

%% figure 3
\begin{figure}[!t]
	\centering
	\includegraphics[width=0.9\linewidth,trim=4 4 4 4,clip]{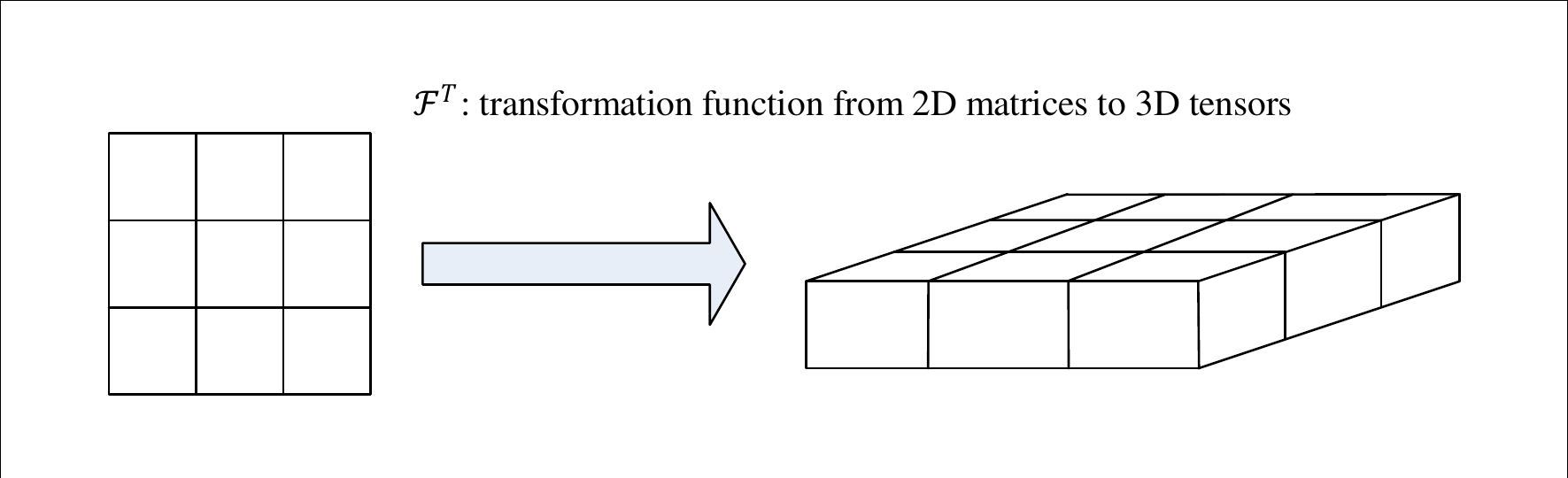}
	\caption{Illustration of the transformation function.}
	\label{figure3}
\end{figure}

\subsection{Fusion convolutional LSTM (FCL-Net)}

In this section, we propose a novel $\textit{fusion convolutional LSTM network}$ (FCL-Net), which integrates spatio-temporary variables and non-spatial time-series variables into one DL architecture for short-term passenger demand forecasting under the on-demand ride service platform. Conv-LSTM layers and convolutional operators are employed to capture characteristics of spatio-temporary variables, while LSTM layers are implemented for non-spatial time-series variables. To fuse these two categories of variables, techniques including repeating and transformation functions, are utilized in the structure. The repeating function is denoted as $\mathcal{F}^{R}(\cdot;M,N):R \to R^{M \times N \times 1}$, where $\left(\mathcal{F}^{R}(x;M,N)\right)_{m,n,1}=x$ for any $m \in (1,2,...,M),n \in (1,2,...,N)$. It is also worth mentioning that the transformation function $\mathcal{F}^{T}$ should be applied to transfer 2D matrices $\bm{D}_t, \bm{\Gamma}_t$ to 3D tensors $\mathcal{D}_t,\mathit{\Gamma}_t \in R^{I \times J \times 1}$ to meet the consistent requirement of convolutional operators.

As mentioned in Problem 1, the forecasting target is the demand intensity during the $t$th time interval, which is denoted as $\mathcal{X}_t=\mathcal{D}_t$.

\subsubsection{Structure for spatio-temporary variables}

Among the variables utilized in this paper, the historical demand intensity and travel time rate are spatio-temporary variables, as denoted in Definition 2. By considering that the historical demand intensity and travel time rate influence the future demand intensity in different ways, these two kinds of variables are fed into two separate architectures, each of which consists of a series of stacked conv-LSTM layers and convolutional operators. Suppose $\textit{K}_d,\textit{K}_\tau$ are the look-back time windows, $\textit{L}_d,\textit{L}_\tau$ are the number of stacked conv-LSTM layers, of the demand intensity and travel time rate, respectively, the formulations of the architecture for spatio-temporary variables are given as follows:

\begin{equation}
(\mathcal{U}_{t-K_d}^{(L_d)},\mathcal{U}_{t-K_d+1}^{(L_d )} ),...,\mathcal{U}_{t-1}^{(L_d)})=\mathcal{F}^{CL}_{L_d}\cdot\cdot\cdot\mathcal{F}^{CL}_{l}\cdot\cdot\cdot\mathcal{F}^{CL}_{1} (\mathcal{D}_{t-K_d},\mathcal{D}_{t-K_d+1},...,\mathcal{D}_{t-1}))
\end{equation}

\begin{equation}
\hat{\mathcal{X}}_t^u=\sigma(\bm{W}_{ux}*\mathcal{U}_{t-1}^{(L_d)}+\bm{b}_u)
\end{equation}

\begin{equation}
(\mathcal{V}_{t-K_\tau}^{(L_\tau)},\mathcal{V}_{t-K_\tau+1}^{(L_\tau)} ),...,\mathcal{V}_{t-1}^{(L_\tau)})=\mathcal{F}^{CL}_{L_\tau}\cdot\cdot\cdot\mathcal{F}^{CL}_{l}\cdot\cdot\cdot\mathcal{F}^{CL}_{1} (\mathit{\Gamma}_{t-K_\tau},\mathit{\Gamma}_{t-K_\tau+1},...,\mathit{\Gamma}_{t-1}))
\end{equation}

\begin{equation}
\hat{\mathcal{X}}_t^v=\sigma(\bm{W}_{vx}*\mathcal{U}_{t-1}^{(L_\tau)}+\bm{b}_v)
\end{equation}
where $\mathcal{U}_{t-k}^{(L_d)},k=1,2,...,K_d, \mathcal{V}_{t-\tau}^{(L_d)},k=1,2,§Ò§ß,K_\tau$ are the output hidden tensors in the highest-level layers of the architectures of demand and travel time rate, respectively. $\bm{W}_{ux},\bm{W}_{vx}$ are convolutional operators utilized to further capture the spatial correlations of the highest-level output tensors, while $\bm{b}_u,\bm{b}_v$ are the intercept parameters. Through these two structures, two high-level components $\hat{\mathcal{X}}_t^u,\hat{\mathcal{X}}_t^v$ can be obtained, which will be further substituted into the fusion layer.

\subsubsection{Structure for non-spatial time-series variables}

Time variables (including the time-of-day and day-of-week) and weather variables (temperature, humidity, weather state, wind speed, and visibility) are the two classes of non-spatial time-series variables. Considering that time variables and weather variables affect the future demand intensity in different ways, we define two sequences of vectors: $\bm{e}_s=(h_s,w_s ),\bm{a}_s=(at_s,ah_s,as_s,aw_s,av_s ),s=1,2,...,t$. These two sequences of vectors are fed into two separate stacked LSTM architectures, which produce the two high-level components $\hat{\mathcal{X}}_t^p$ and $\hat{\mathcal{X}}_t^q$.

\begin{equation}
(\bm{p}_{t-K_e+1}^{(L_e)},...,\bm{p}_{t}^{(L_e)})=\mathcal{F}^{L}_{L_e}\cdot\cdot\cdot\mathcal{F}^{L}_{l}\cdot\cdot\cdot\mathcal{F}^{L}_{1} (\bm{e}_{t-K_e+1},...,\bm{e}_{t-1},\bm{e}_{t})
\end{equation}

\begin{equation}
\hat{\mathcal{X}}_t^p=\mathcal{F}^{T}(\mathcal{F}^{R}(\sigma(\bm{w}_{p}\bm{p}_{t}^{(L_e)}+{b}_p)))
\end{equation}

\begin{equation}
(\bm{q}_{t-K_a+1}^{(L_a)},...,\bm{q}_{t-1}^{(L_a)})=\mathcal{F}^{L}_{L_a}\cdot\cdot\cdot\mathcal{F}^{L}_{l}\cdot\cdot\cdot\mathcal{F}^{L}_{1} (\bm{a}_{t-K_a},...,,\bm{a}_{t-1})
\end{equation}

\begin{equation}
\hat{\mathcal{X}}_t^q=\mathcal{F}^{T}(\mathcal{F}^{R}(\sigma(\bm{w}_{p}\bm{q}_{t-1}^{(L_a)}+{b}_q)))
\end{equation}
where $\bm{p}_{t-k}^{(L_h)},k=1,2,...,K_e,\bm{q}_{t-K}^{(L_a)}),k=1,2,§Ò§ß,K_a$ are the output hidden vectors in the highest LSTM layers $L_e,L_a$.

\subsubsection{Fusion}
Inspired by the fact that the high-level components have different contributions to the prediction, we employ Hadamard product `$\circ$' to multiply these components by the four parameter matrices $\bm{W}_u,\bm{W}_v,\bm{W}_p$ and $\bm{W}_q$, which can be learnt to evaluate the importance of the components during the training process. Therefore, the estimated demand intensity during the  $t$th time interval is given by

\begin{equation}
\hat{\mathcal{X}}_t=\bm{W}_u\circ\hat{\mathcal{X}}_t^u+\bm{W}_v\circ\hat{\mathcal{X}}_t^v+\bm{W}_p\circ\hat{\mathcal{X}}_t^p+\bm{W}_q\circ\hat{\mathcal{X}}_t^q
\end{equation}

\subsubsection{Objective function}
During the training process of the FCL-Net, the object is to minimize the mean squared error between the estimated and real demand intensity, through which the weighted and intercept parameters can be learnt. The objective function of the architecture is shown in Eq. (22).

\begin{equation}
\min_{w,b} \left \| {\mathcal{X}}_t - \hat{\mathcal{X}}_t \right \| _2^2 + \alpha \left\| \bm{W} \right\| _2^2
\end{equation}

The second term of the objective function represent an L2 norm regularization term, which helps avoid over-fitting issues. $\bm{W}$ stands for all the weighted parameters in $\hat{\mathcal{X}}_t$, and $\alpha$ refers to a regularization parameter which balances the bias-variance tradeoff.

The training steps of the FCL-Net is illustrated in Algorithm 1.

\begin{algorithm}[H]
	\caption{FCL-Net Training}\label{euclid}
	\begin{tabularx}{\textwidth}{lX}
		\textbf{Input} & Observations of demand intensity $\{\bm{D}_1,...,\bm{D}_n\}$  in training dataset\\
		& Observations of demand intensity $\{\bm{\Gamma}_1,...,\bm{\Gamma}_n\}$  in training dataset\\
		& Observations of time-of-day $\{h_1,...,h_n\}$, day-of-week $\{w_1,...,w_n\}$ in training dataset\\
		& Observations of weather variables $\{at_1,...,at_n\}$, $\{ah_1,...,ah_n\}$, $\{as_1,...,as_n\}$, $\{aw_1,...,aw_n\}$, $\{av_1,...,av_n\}$\\
		& lookback-windows: $K_d$,$K_\tau$,$K_e$,$K_a$
	\end{tabularx}
	
	\begin{tabularx}{\textwidth}{lX}
		\textbf{Output} & FCL-Net with learnt parameters\\
	\end{tabularx}

	\begin{algorithmic}[1]
				
		\Procedure{FCL-Net Training}{}
		\State Initialize a null set: $L \gets \emptyset$
		\For {all available time intervals $t$ $(1 \leqslant t \leqslant n)$}
			\State $\mathcal{S}_t^d \gets [\mathcal{D}_{t-K_d},\mathcal{D}_{t-K_d+1},...,\mathcal{D}_{t-1}]$
			\State $\mathcal{S}_t^{\tau} \gets [\mathit{\Gamma}_{t-K_{\tau}},\mathit{\Gamma}_{t-K_{\tau}+1},...,\mathit{\Gamma}_{t-1}]$
			\State $\mathcal{S}_t^e \gets [\bm{e}_{t-K_e},...,\bm{e}_{t-1},\bm{e}_{t}]$, where $\bm{e}_s=(h_s,w_s)$
			\State $\mathcal{S}_t^a \gets [\bm{a}_{t-K_a},...,\bm{a}_{t-1},\bm{a}_{t}]$, where $\bm{a}_s=(at_s,ah_s,as_s,aw_s,av_s)$
			\Comment{where $\mathcal{S}_t^d$, $\mathcal{S}_t^{\tau}$,$\mathcal{S}_t^e$,$\mathcal{S}_t^a$ are the sets of different categories of explanatory variables in one observation.}
			\State A training observation $(\{\mathcal{S}_t^d, \mathcal{S}_t^{\tau},\mathcal{S}_t^e,\mathcal{S}_t^a\},\mathcal{D}_t)$ is put into $L$
		\EndFor
		\State Initialize all the weighted and intercept parameters
		\Repeat
			\State Randomly extract a batch of samples $L^b$ from $L$
			\State Estimate the parameters by the minimizing the objective function shown in Eq. (22) within $L^b$
		\Until {convergence criterion met}
		\EndProcedure
	\end{algorithmic}
\end{algorithm}

%% figure 4
\begin{figure}[!t]
	\centering
	\includegraphics[width=0.9\linewidth,trim=4 4 4 4,clip]{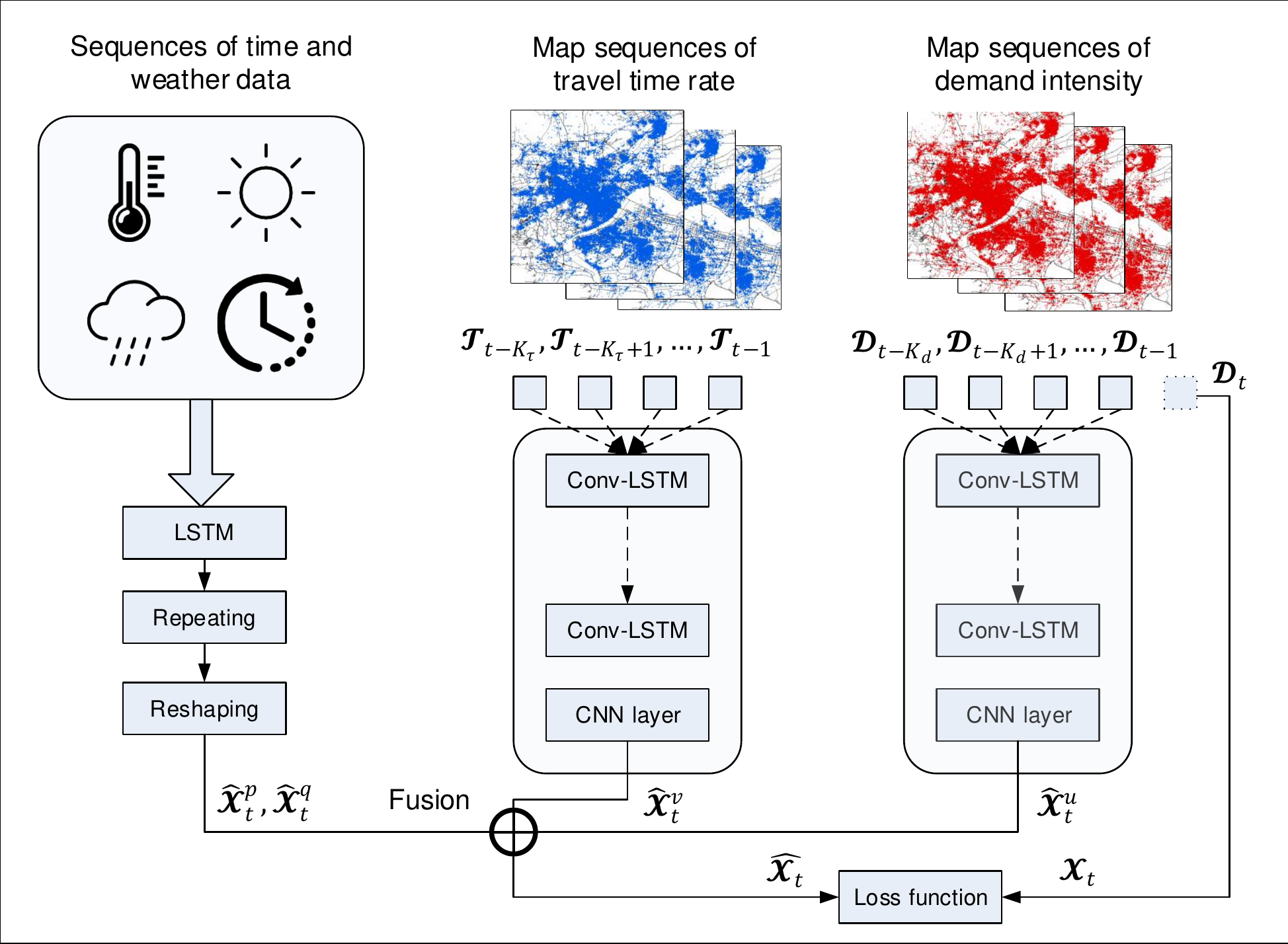}
	\caption{Framework of the proposed FCL-Net approach.}
	\label{figure4}
\end{figure}

\subsection{Spatial aggregated random forest for feature selection}

Random forest, first introduced by \cite{breiman2001random}, is one of the most powerful ensemble learning algorithms for regression problems. Consider a training set with $m$ observations, $L=\{(\bm{X}^{(1) },y^{(1)} ),...,(\bm{X}^{(m) },y^{(m)} )\}$, where $\bm{X}^{(i)} \in R^p $ is the $i$th observation of features, and $y^{(i)} \in R$ is the $i$th observation of label. A random forest builds $K$ decision trees by generating $K$ sets of bootstrap sample sets, $L_1,...,L_K$, from the training set $L$, while the $k$th decision tree can be represented as $\textit{f}_k:R^p \to R$. The out-of-bag error of the $k$th tree, $errO_k$, is denoted as the average error in the out-of-bag sample sets, $O_1,...,O_K$, with respect to each tree (shown in Eq. (23)).

\begin{equation}
\textit{errO}_k = \frac{1}{n}\mathop  \sum \limits_{i \in O_k} {\left( {y^{(i)} - \hat{y}^{(i)}} \right)}^2
\end{equation}
where $\hat{y}^{(i)}=f_k \left( \bm{X}^{(i)} \right) $ is the estimated value of the $i$th labels based on tree $\textit{k}$. The out-of-bag error can be utilized to calculate the feature importance through the following steps \citep{genuer2015vsurf}: (1) permute the $j$th variable of $\bm{X}$ in each $O_k$ to get a new out of bag samples ${O_k}^{'}$; (2) calculate the out-of-bag error, $ \widetilde{errO}_k^j$ in the new sets of samples ${O_k}^{'}$; (3) the importance of the $j^{th}$ variable, $ VI \left( \textit{X}_j \right) $, is equal to the average difference between $errO_k$ and $\widetilde{errO}_k^j$ of all trees (shown in Eq. (24)).

\begin{equation}
\textit{VI} \left( \textit{X}_j \right) = \frac{1}{\textit{K}} \mathop \sum \limits_k {\left( \widetilde{errO}_k^j - errO_k \right)}
\end{equation}

Considering that the dependent variable in the passenger demand forecasting, $\bm{D}_t  \in R^{I \times J}$, is an ${I \times J}$ matrix, instead of a continuous value in the standard random forest, we develop a spatial aggregated random forest which consists of ${I \times J}$ standard random forests, to examine the aggregated variable importance partitioned by category and look-back time window. To illustrate the spatial aggregated random forest, we extend Problem 1 to Problem 2, given by

\textbf{Problem 2}: Given the historical observations and known information $\{d_s^{i,j},\tau_s^{i,j} |s=t-K,...,t-1, i \in {1,§Ò§ß,I},j \in \{1,...,J\}; h_s,w_s |s=t-K+1,...,t; {at}_s,{ah}_s,{as}_s,{aw}_s,{av}_s |s=t-K,...,t-1\}$, predict $d_s^{i{'},j{'}}$ via the standard random forest $f^{i{'},j{'}}$, for all $i^{'} \in \{1,...,I\},j^{'} \in \{1,...,J\}$. The length of look-back window is denoted as K.

$VI^{i{'},j{'}}$ is denoted as the function to calculate variable importance in random forest $f^{i{'},j{'}}$. Two tensors, $\mathbb{V}^d, \mathbb{V}^\tau \in R^{I \times J \times I \times J \times K}$, where $\left(\mathbb{V}^d \right)_{i,j,i^{'},j^{'},k}=VI^{i{'},j{'}} \left(d_{t-k}^{i,j} \right), \left(\mathbb{V}^\tau \right)_{i,j,i^{'},j^{'},k}=VI^{i{'},j{'}} \left(\tau_{t-k}^{i,j} \right)$, are denoted to store the variable importance of two categories of spatial-temporal variables. $\left(\mathbb{V}^d \right)_{i,j,i^{'},j^{'},k}$ refers to the variable importance of the passenger demand in $\{i,j\}$ during time interval t-k in the problem of forecasting passenger demand of $\{i^{'},j^{'}\}$ during time slot t. As for non-spatial time-series variables, we define $\mathbb{V}^h,\mathbb{V}^w ,\mathbb{V}^{at} ,\mathbb{V}^{ah},\mathbb{V}^{as},\mathbb{V}^{aw} ,\mathbb{V}^{av} \in R^{I \times J \times K}$, where $\left(\mathbb{V}^h \right)_{i,j,i^{'},j^{'},k}=VI^{i{'},j{'}} \left(h_{t-k} \right)$, and the same expression for $w,at,ah,as,aw,av$. All the variable importance in  $\mathbb{V}^d,\mathbb{V}^\tau,\mathbb{V}^h,\mathbb{V}^w,\mathbb{V}^{at},\mathbb{V}^{ah},\mathbb{V}^{as},\mathbb{V}^{aw} ,\mathbb{V}^{av}$ is normalized to percentage via dividing each variable importance by the sum of all variable importance.

Firstly, we examine the variable importance partitioned by category, i.e. $\sum_i \sum_j \sum_{i^{'}} \sum_{j^{'}} \sum_k \mathbb{V}^d ,\sum_{i^{'}} \sum_{j^{'}}\sum_k \mathbb{V}^h$, etc., to select the important variables in terms of category. Secondly, we investigate the variable importance partitioned by category and look-back time window $k (k \in \{1,2,...,K\})$, to select a suitable look-back window for each category of variable.

\section{Experiments and Results}

\subsection{On-demand ride service platform data}

The datasets utilized in this paper are extracted from DiDi Chuxing, the largest on-demand ride service platform in China, during one-year period between November 1, 2015 and November 1, 2016. We randomly obtain 1,000,000 requesting orders from the platform, each of which consists of the requesting time, travel distance, travel time, longitude and latitude. The study site is located in Hangzhou, China, starting from 120.00 to 120.35 in longitude, and from 30.45 to 30.15 in latitude. The dataset is partitioned into 1-hour time intervals, and the investigated region is partitioned into $7 \times 7$ grids, as shown in Fig. 5. The one-hour aggregated weather variables, including temperature, humidity, weather state, wind speed, and visibility, are obtained during the same period.

To avoid using future information, the dataset is divided into 70\% training dataset comprised of observations between November 1, 2015 and July 14, 2016, and the 30\% test dataset consisting of the remaining observations between July 15, 2016 and November 1, 2016.

%% figure 5
\begin{figure}[!t]
	\centering
	\includegraphics[width=0.6\linewidth,trim=4 4 4 4,clip]{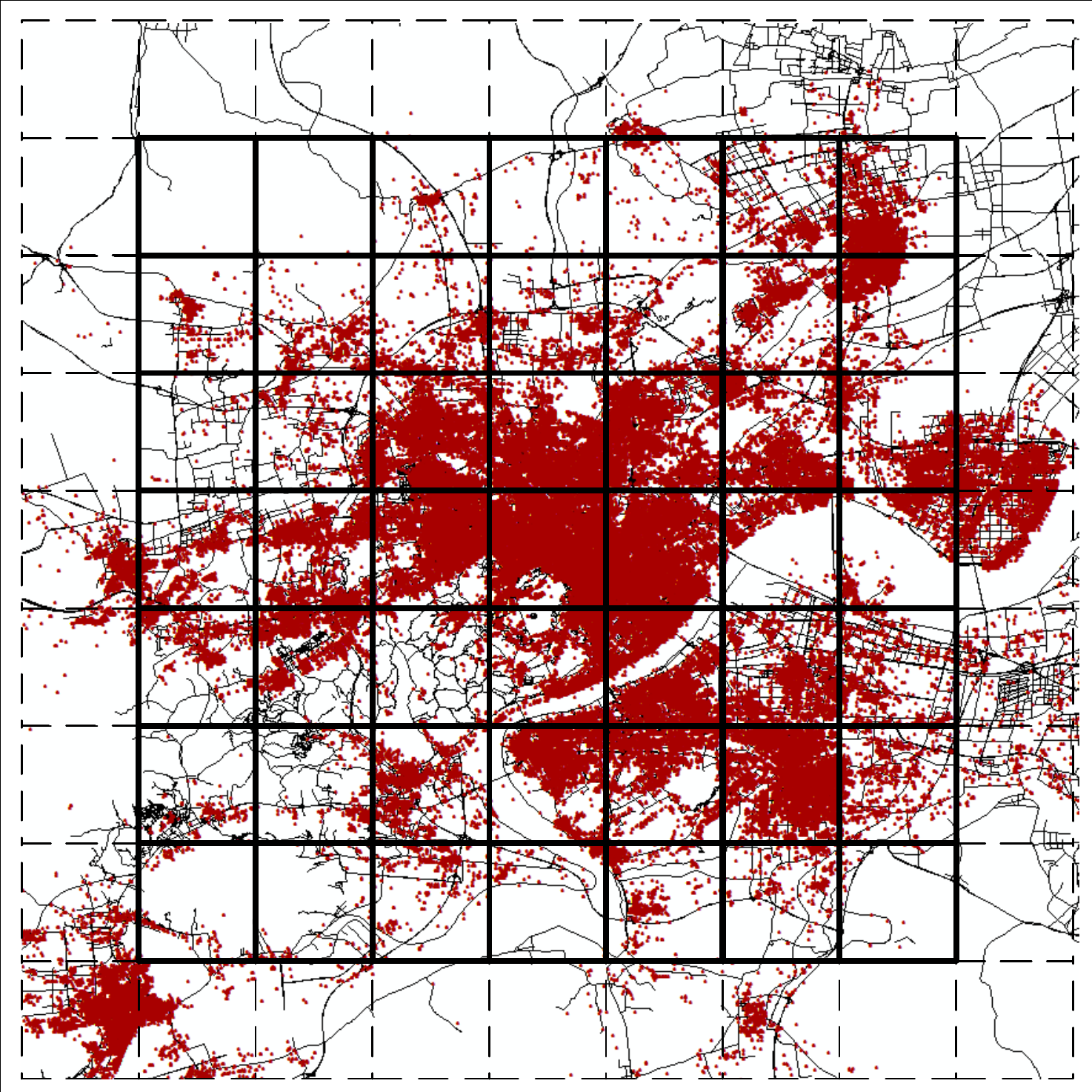}
	\caption{The investigated region partitioned into $7\times7$ grids.}
	\label{figure5}
\end{figure}

%% figure 6
\begin{figure}[!t]
	\centering
	\includegraphics[width=0.6\linewidth]{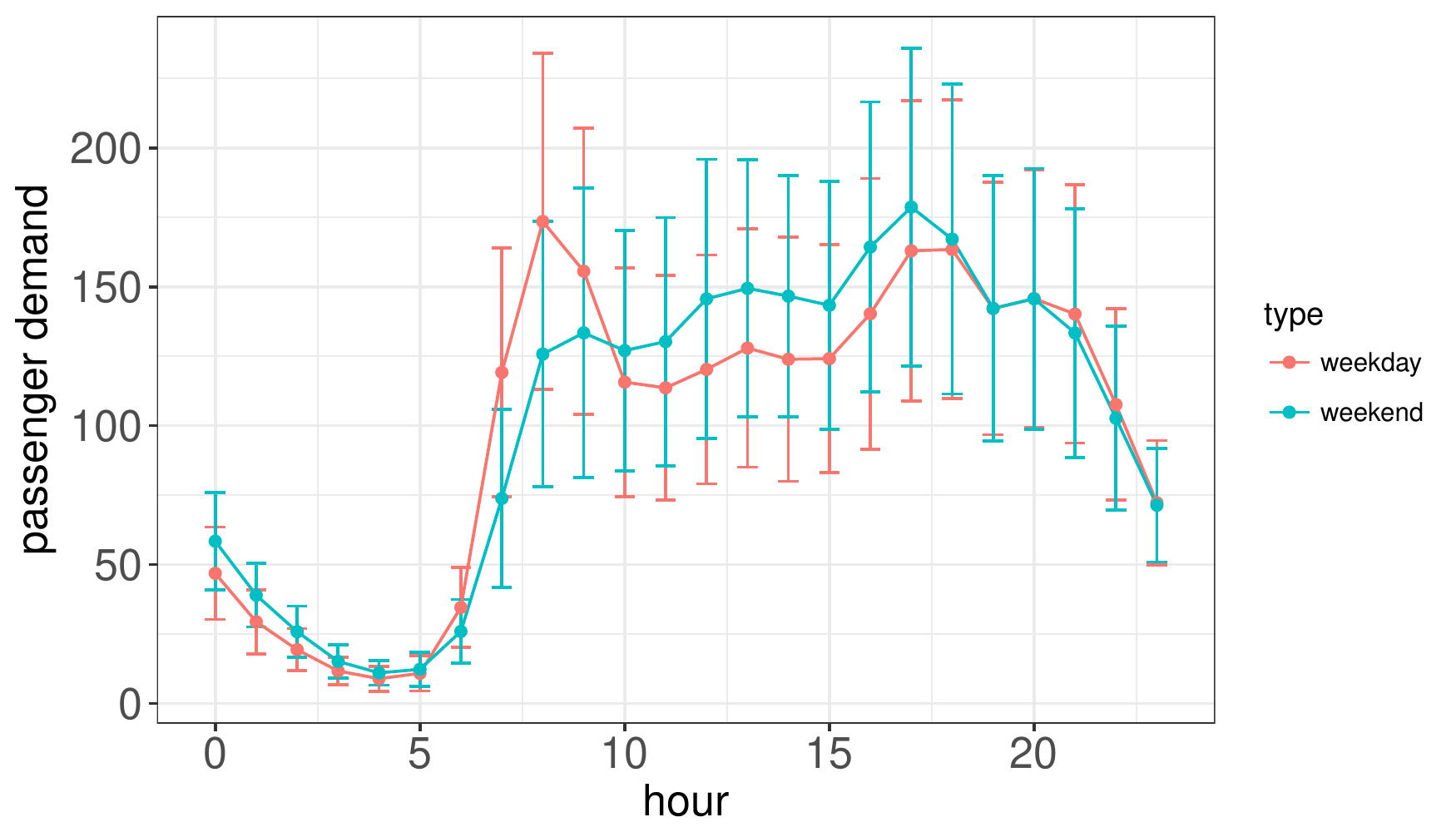}
	\caption{Mean and standard deviation of passenger demand in different hours.}
	\label{figure6}
\end{figure}

It can be observed from Fig. 6, which shows the mean and variance of passenger demand in different hours of a day based on the training dataset, that the passenger demand in weekdays demonstrates a double-peak nature while the passenger demand in weekends shows a single-peak property. Therefore, the peak hours, off-peak hours, and sleep hours are separately defined for weekdays and weekends, in this paper.

\subsubsection{Exploring the spatio-temporal correlations}

The reason for utilizing a tailored spatio-temporal DL architecture is that there exist spatio-temporal correlations among the spatio-temporal variables, i.e., the demand intensity and travel time rate. To validate this assumption, we examine the correlations between demand intensity at the $t$th time interval and spatio-temporal variables ahead of the $t$th time interval by employing the Pearson correlation, given by

\begin{equation}
\textnormal{Corr} (\bm{Y},\bm{Z}) = \dfrac{E[\left(\bm{Y}-E(\bm{Y})\right)^{'} \left(\bm{Z}-E(\bm{Z})\right)]}{E[(\bm{Y}-E\left(\bm{Y})\right)^{2}] E[(\bm{Z}-E\left(\bm{Z})\right)^{2}]}
\end{equation}
where $\bm{Y,Z}$ are two random variables with the same number of observations.

Firstly, we calculate the Pearson correlations between the demand intensity at time $t$ in grid $(i^{'},j^{'})$ and demand intensity, travel time rate at $t-k$ time interval in grid $(i,j)$, for all $i,i^{'} \in \{1,...,I\},j,j^{'}  \in \{1,...,J\},k \in \{1,2,3,4\}$. Secondly, we average these correlations partitioned by spatial distances and look-back time intervals. The spatial distance of grid $(i,j)$ and $(i^{'},j^{'})$ is denoted as the Euclidean distance between the central points of two grids.

Fig. 7 shows the average correlations between the dependent variable (demand intensity at time $t$ in grid $(i^{'},j^{'})$) and the explanatory variables (demand intensity and travel time rate at time $t-k$ in grid $(i,j))$. It can be observed that the average correlations drop gradually not sharply, with the increase of spatial distance, indicating that there exit strong spatial correlations among each grid and its neighbors. On the other hand, it is not surprising that variables with shorter look-back time intervals have higher correlations, but the variables with large look-back time intervals are also correlated with the to-be-predicted demand intensity to some extent. This correlation analysis of the dataset provides strong evidence of the spatial and temporal dependences existing among the spatio-temporal variables.

%% figure 7
\begin{figure}[!t]
	\centering
	\subfloat[demand intensity]{\includegraphics[width=0.45\linewidth]{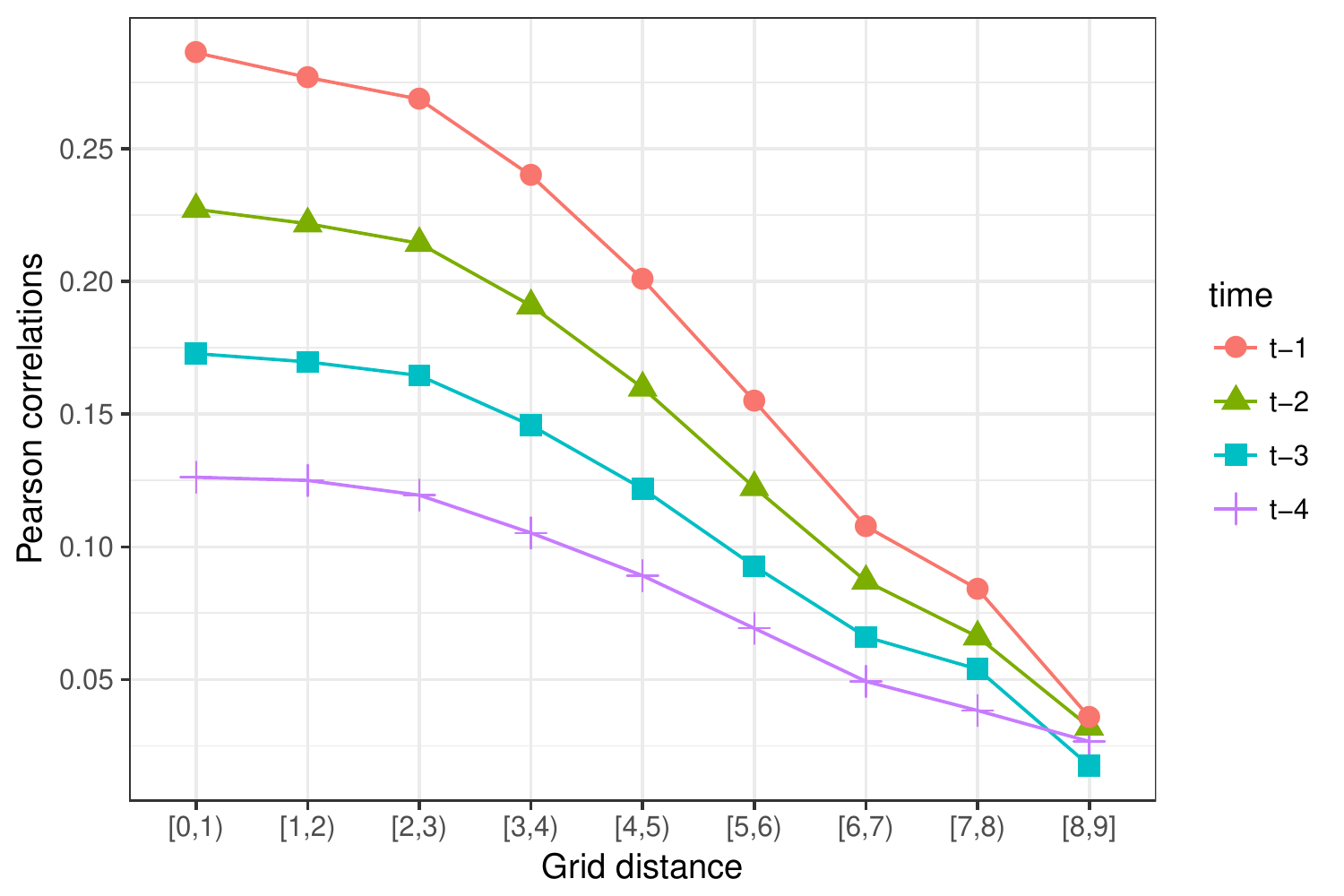}%
		\label{figure7a}}
	\hfil
	\subfloat[travel time rate]{\includegraphics[width=0.45\linewidth]{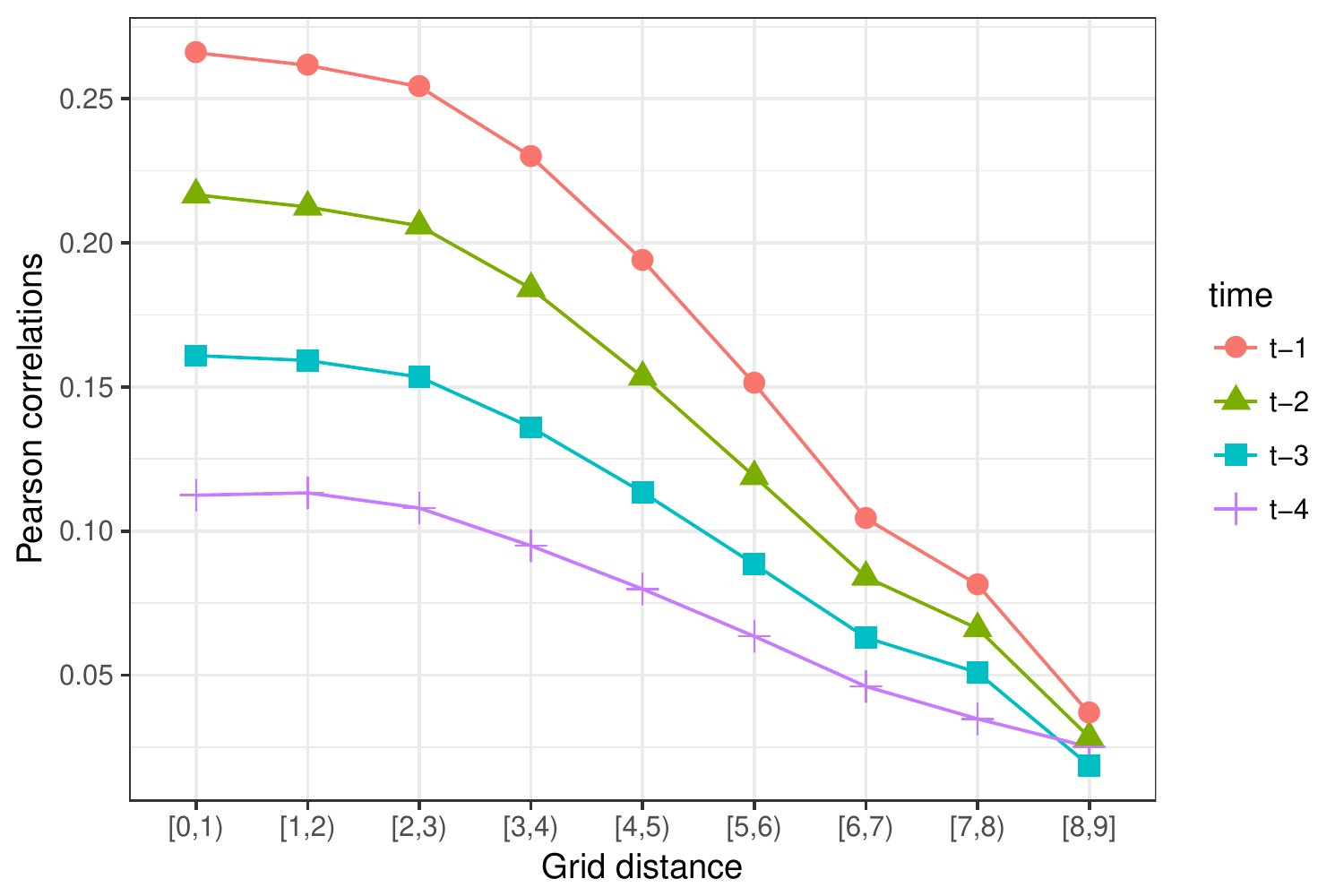}%
		\label{figure7b}}
	\caption{Average correlations partitioned by distance and time.}
	\label{figure7}
\end{figure}

\subsection{Feature selection}

Firstly, Fig. 8 shows the variable importance partitioned by category, based on our proposed spatial aggregated random forest algorithm. It can be observed that the two categories of spatial-temporary variables, travel time rate and demand intensity, are the dominating factors, followed by time-of-day and temperature. However, other variables, such as day-of-week, humidity, etc., have little contributions (less than 5\%) to the prediction.

%% figure 8
\begin{figure}[!t]
	\centering
	\includegraphics[width=0.7\linewidth]{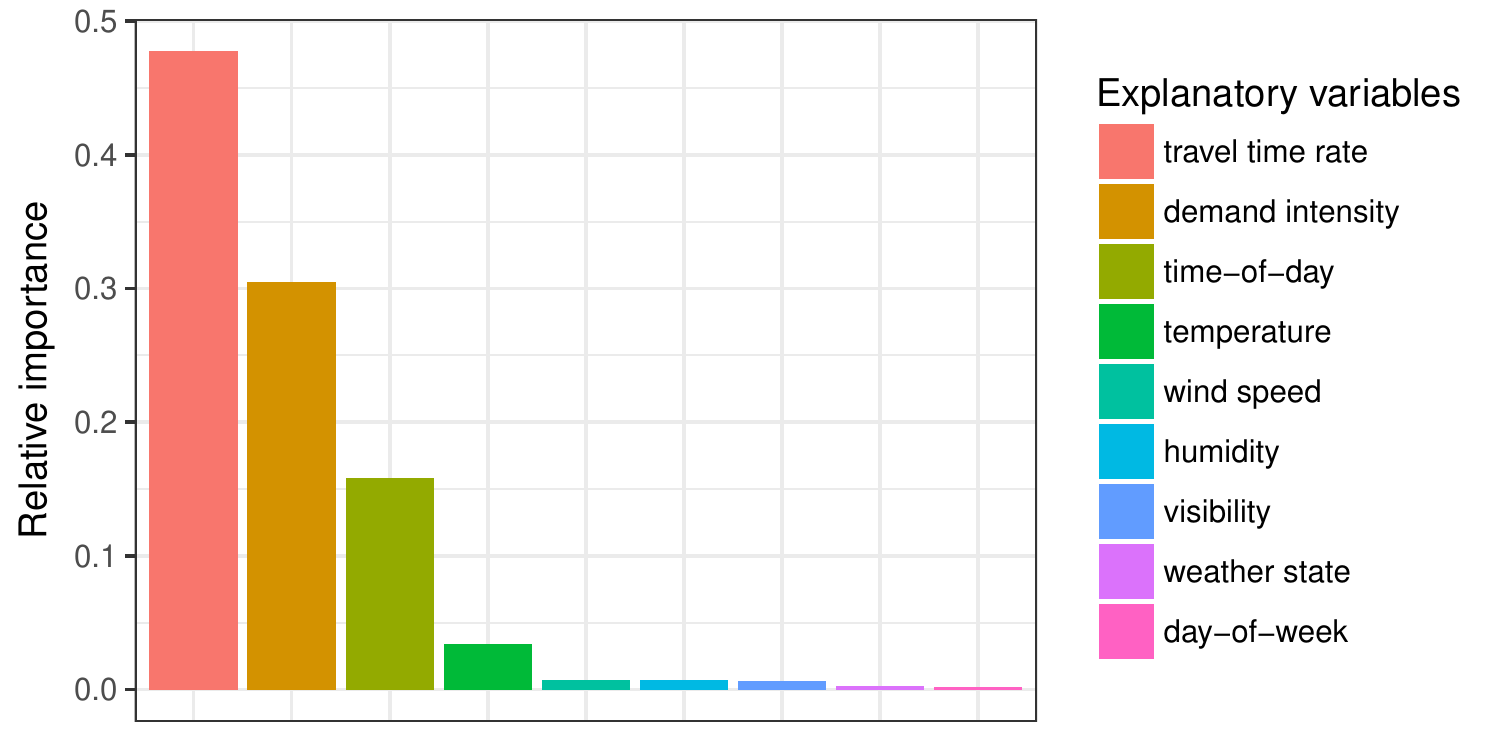}
	\caption{Variable importance partitioned by category.}
	\label{figure8}
\end{figure}

Secondly, Table 1 presents the relative importance of the variables sorted by category and look-back time interval (``-'' represents not applicable). Fig. 9 displays the top 20 important variables. We set the look-back time window $K$ to be 8 for all category of variables. It can be found that the time-of-day during time interval $t$ is the most important variable, followed by the demand intensity and travel time rate during $t-1$ time interval. The importance of all kinds of variables decreases with the look-back time window, but different categories of variables show different descent speeds. The travel time rate far before $t$ still has considerable variable importance, while the time-of-day prior to time $t$ makes little contribution.

\begin{table}[!t]
	\caption{The relative importance of variables partitioned by category and look-back time interval (${\%}$)}
	\small
	\centering
	\begin{tabularx}{1\textwidth}{cccccccccccc}
		\toprule
		\toprule
		${s}$ & ${D_s}$  & ${\Gamma_s}$ & ${at_s}$ & ${ah_s}$ & ${as_s}$ & ${aw_s}$ & ${av_s}$ & ${h_s}$ & ${w_s}$\\
		\midrule
		$t-8$ & 2.628 & 5.005 & 0.589 & 0.074 & 0.036 & 0.080 & 0.089 & - & - \\
		$t-7$ & 2.040 & 4.432 & 0.275 & 0.084 & 0.035 & 0.086 & 0.076 & 0.137 & 0.041 \\
		$t-6$ & 1.978 & 4.790 & 0.251 & 0.088 & 0.032 & 0.083 & 0.072 & 0.039 & 0.018 \\
		$t-5$ & 2.280 & 4.852 & 0.223 & 0.071 & 0.032 & 0.074 & 0.091 & 0.013 & 0.013 \\
		$t-4$ & 2.895 & 4.984 & 0.205 & 0.085 & 0.043 & 0.079 & 0.089 & 0.021 & 0.022 \\
		$t-3$ & 3.140 & 6.038 & 0.333 & 0.097 & 0.033 & 0.069 & 0.071 & 0.514 & 0.013 \\
		$t-2$ & 3.518 & 7.008 & 0.450 & 0.110 & 0.039 & 0.172 & 0.070 & 0.589 & 0.022 \\
		$t-1$ & 11.999 & 10.658 & 1.105 & 0.116 & 0.039 & 0.083 & 0.076 & 0.671 & 0.015 \\
		$t$ & - & - & - & - & - & - & - & 13.808 & 0.012 \\
		\bottomrule
		\bottomrule
	\end{tabularx}
\end{table}

Selecting appropriate categories of variables and the suitable look-back window helps to improve computation efficiency with little loss of predictive performance. In this paper, by considering the trade-off between computation efficiency and predictive performance, we select 4 categories of variables: demand intensity, travel time rate, time-of-day, and temperature, with 4,8,2,2 look-back time windows, respectively. This feature selection reduces the number of variables in each observation from $8§Ò©Ø7§Ò©Ø7§Ò©Ø2+8§Ò©Ø5+8§Ò©Ø2=840$ to $8§Ò©Ø7§Ò©Ø7+4§Ò©Ø7§Ò©Ø7+2+2=592$, which indicates 29.5\% decrease in computational complexity.

%% figure 9
\begin{figure}[!t]
	\centering
	\includegraphics[width=1.0\linewidth]{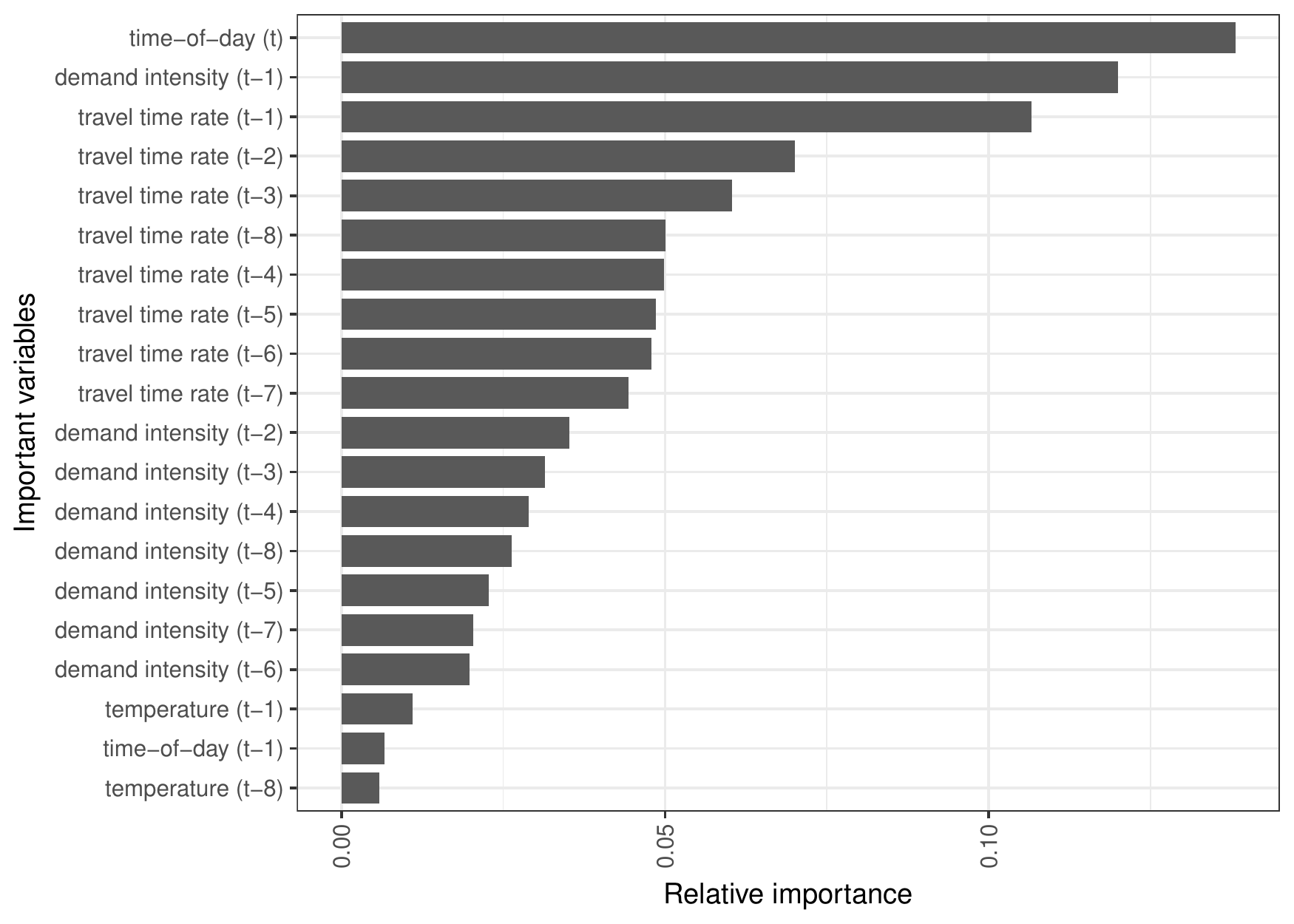}
	\caption{Top 20 important variables partitioned by category and time.}
	\label{figure9}
\end{figure}

\subsection{Model comparisons}

The proposed FCL-Net with full variables and selected variables (selected by the spatial aggregated random forest) are trained on the training dataset and validated on the test dataset, respectively. Meanwhile, the conv-LSTM network, which is only fed with the historical demand intensity, is trained and tested in the same way. The definition of the above-mentioned three models are shown as follows:

\begin{enumerate}[(1)]
  \item \textit{Conv-LSTM with only historical demand intensity}: 	this model utilizes historical observations of demand intensity $\{\bm{D}_s |s=t-K_d,...,t-1\}$ to predict future demand intensity $\bm{D}_t$, where $K_d$=8 in this paper. The architecture of Conv-LSTM is introduced in Section 4.2;

  \item \textit{FCL-Net with full variables}: historical observations of all variables $\{\bm{D}_s,{\Gamma}_s | s=t-K,...,t-1;h_s,w_s |s=t-K+1,...,t;{at}_s,{ah}_s,{as}_s,{aw}_s,{av}_s |s=t-K,...,t-1 \}$, where$ K=8$, are utilized to predict future demand intensity $\bm{D}_t$. The training process of this model is illustrated in Algorithm 1;

  \item \textit{FCL-Net with selected variables}: this model utilizes the historical observations of the selected variables $\{\bm{D}_s |s=t-K_d,...,t-1,\Gamma_s |s=t-K_\tau,...,t-1;h_s |s=t-K_h,...,t;at_s |s=t-K_{at},...,t-1\}$ where $K_d=4,K_\tau=8,K_h=2,K_{at}=2$, to forecast future demand intensity $\bm{D}_t$. The training process of this model is the same as Algorithm 1 except that the inputs are replaced by the selected variables.
\end{enumerate}

Apart from the proposed three models, several benchmark algorithms are also tested. The benchmark algorithms include three traditional time-series forecasting models (HA, MA, and ARIMA) and two classical neural networks (ANN and LSTM).

\begin{enumerate}[(1)]
	\item \textit{HA}: 	the historical average model predicts future demand intensity in the test dataset based on the empirical statistics in the training dataset. For example, the average demand intensity during 8-9 AM in grid $(i,j)$ is estimated by the mean of all historical demand intensity during 8-9 AM in grid $(i,j)$.
	
	\item \textit{MA}: 	the moving average model is widely-used in time-series analysis, which predicts future value by the mean of serval nearest historical values. In this paper, the average of 8 previous demand intensity in grid $(i,j)$ is used to predict the future demand intensity in grid $(i,j)$.
	
	\item \textit{ARIMA}: 	the autoregressive integrated moving average model indicates the autoregressive (AR), integrated (I), and MA parts, and the model considers trends, cycles, and non-stationary characteristics of a dataset simultaneously.
	
	\item \textit{ANN}: the artificial neural network employs all the variables, including historical demand intensity, travel time rate, hour and week state, and weather variables, with look-back time window $K=8$, of a specific grid $(i,j)$, to predict the future demand intensity in grid $(i,j)$. ANN does not differentiate variables across time and thus fails to capture time dependences.
	
	\item \textit{LSTM}: The LSTM utilizes all the variables, including historical demand intensity, travel time rate, hour and week state, and weather variables, with look-back time window $K=8$, of grid $(i,j)$, to predict future demand intensity in grid $(i,j)$. Unlike ANN, LSTM considers temporal dependences, but does not capture spatial dependences.
\end{enumerate}

We evaluate the models via three effectiveness of measures: root mean squared error (RMSE), coefficient of determination $(\rm{R}^2)$, and mean absolute error (MAE), the formulation of which are given as follows.

\begin{equation}
{\rm{RMSE}} = \sqrt {\frac{1}{n}\mathop \sum \limits_{i=1} {{\left( {y^{(i)} - \hat{y}^{(i)}} \right)}^2}}
\end{equation}

\begin{equation}
{\rm{R}^2} = 1- \dfrac{\sum \limits_{i=1}{{\left( {y^{(i)} - \hat{y}^{(i)}} \right)}^2}}{\sum \limits_{i=1}{{\left( y^{(i)} - \bar{y} \right)}^2}}
\end{equation}	

\begin{equation}
{\rm{MAE}} = \frac{1}{n}\mathop \sum \limits_{i=1} { | {y^{(i)} - \hat{y}^{(i)}} | ^2}
\end{equation}
where $y^{(i)}$, $\hat{y}^{(i)}$ are the $i$th ground truth and estimated value of demand intensity, $\bar{y}$ is the mean of all $y^{(i)}$, and $n$ is the size of the test set.

Before model training and validation, the demand intensity, travel time rate, hour-of-day and day-of-week, and weather variables, are standardized to the range [0,1], through the max-min standardization, respectively.

Table 2 shows the predictive performance comparison results of the proposed models and benchmark models on the test set. It can be found that the proposed FCL-Net outperforms other methods.

Both FCL-Nets have relatively 50.9\% lower RMSE than Conv-LSTM with only historical demand intensity, which indicates that the exogenous variables make great contribution to the short-term passenger demand forecasting. As mentioned above, the proposed spatial aggregated random forest reduces the computation complexity of FCL-Net by 29.5\% (the number of variables of in each observation drops from 840 to 592). Meanwhile, Table 2 shows that FCL-Net only suffers a 0.6\% decrease measured by RMSE, 0.1\% decrease by $\textnormal{R}^2$, or 1.1\% decrease by MAE, on predictive performance after feature selection. The results indicate that the feature selection process is valuable to FCL-Net since it balances the computation complexity and predictive performance.

Figure 10 shows some samples of heat maps of the ground truth passenger demand and predicted results by FCL-Net, where the deeper color implies a larger demand intensity. It is obvious that the demand intensity in peak hours (e.g., 9-10 AM and 6-7 PM) is much higher than that in sleep hours (e.g., 0-1 AM). The demand intensity is unbalanced across space: the central grids have much a higher demand intensity than other grids. The trend of the demand intensity over time is even different in different grids and different days, which makes it hard to forecast short-term passenger demand. From the samples of visualization, we can find that the FCL-Net primarily captures the spatio-temporal characteristics of the demand intensity and makes more accurate forecasting. The combination of short-term passenger demand forecasting and visualization helps traffic operators of the platform/government to detect and forecast grids with oversupply and overfull demand and design proactive strategies to avoid these imbalanced conditions.

\begin{table}[!t]
	\caption{Predictive performance comparison}
	\centering
	\begin{tabularx}{1\textwidth}{cccc}
		\toprule
		\toprule
		${\bold{Model}}$ & ${\bold{RMSE}}$  & ${R^2}$ & ${\bold{MAE}}$\\
		\midrule
		HA & 0.0378 & 0.736 & 0.0192\\
		MA & 0.0511 & 0.518 & 0.0260\\
		ARIMA & 0.0345 & 0.780 & 0.0178\\
		ANN & 0.0331 & 0.798 & 0.0194\\
		LSTM & 0.0322 & 0.808 & 0.0181\\
		Conv-LSTM (with only demand intensity) & 0.0318 & 0.813 & 0.0176\\
		FCL-Net (with full variables) & 0.0156 & 0.820 & 0.0090\\
		FCL-Net (with selected variables) & 0.0157 & 0.819 & 0.0091\\
		\bottomrule
		\bottomrule
	\end{tabularx}
\end{table}

%% figure 10
\begin{figure}[!t]
	\centering
	\subfloat[0-1 AM (G)]{\includegraphics[width=0.23\linewidth]{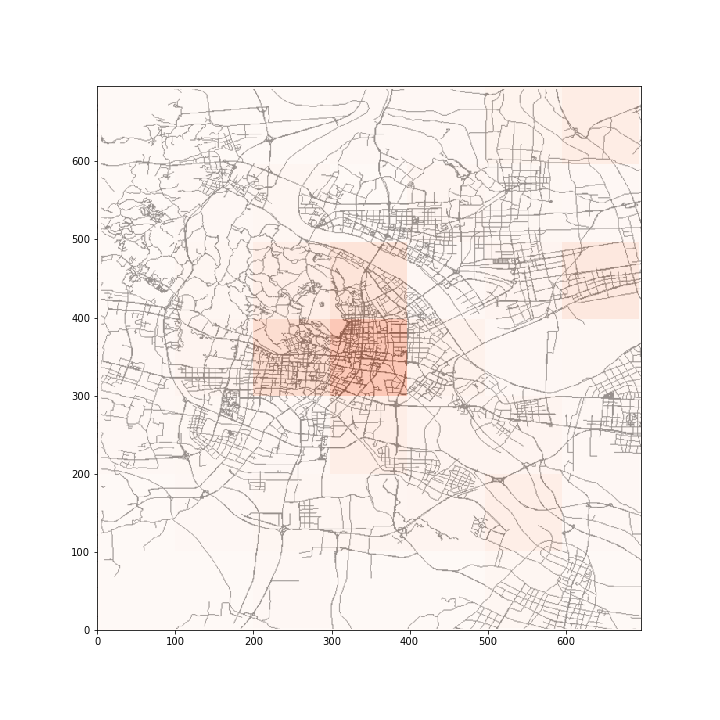}%
		\label{figure10a}}
	\hfil
	\subfloat[9-10 AM (G)]{\includegraphics[width=0.23\linewidth]{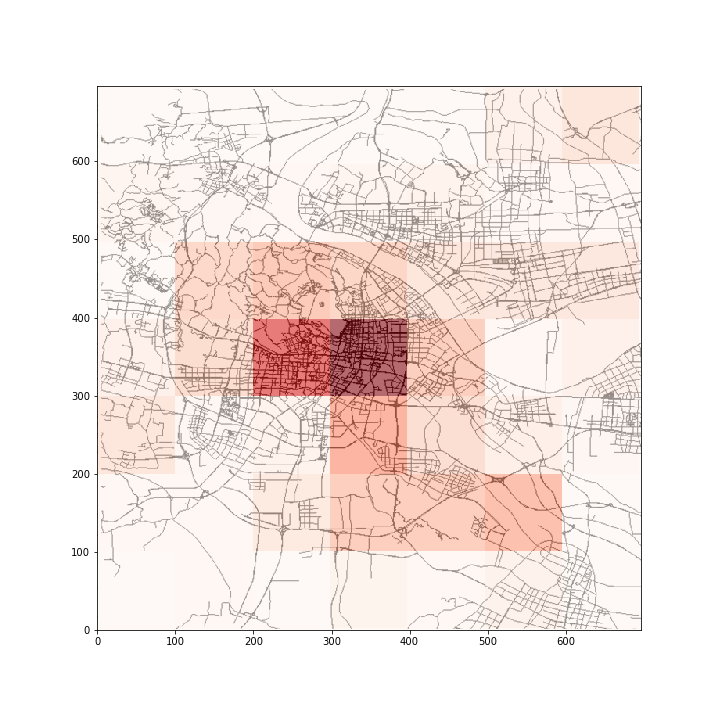}%
		\label{figure10b}}
	\hfil
	\subfloat[6-7 AM (G)]{\includegraphics[width=0.23\linewidth]{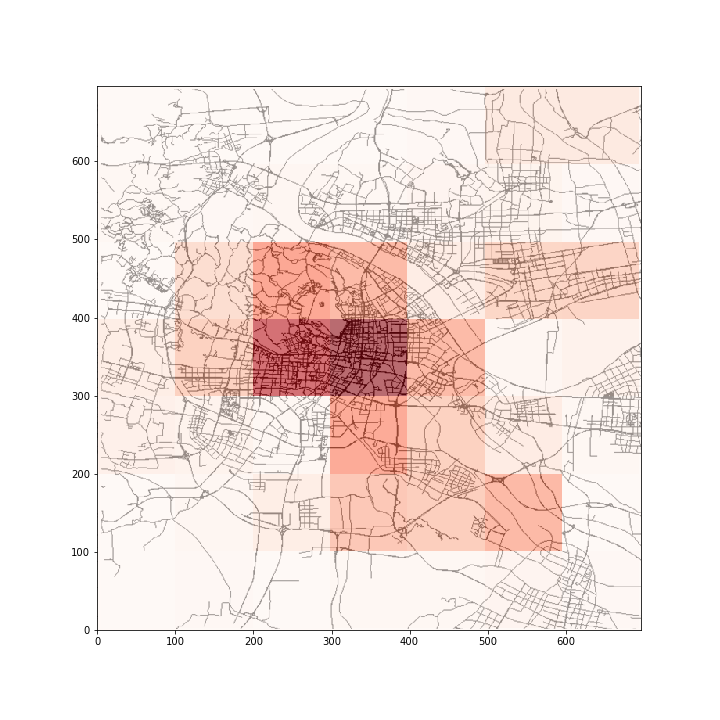}%
		\label{figure10c}}
	\hfil
	\subfloat[9-10 PM (G)]{\includegraphics[width=0.23\linewidth]{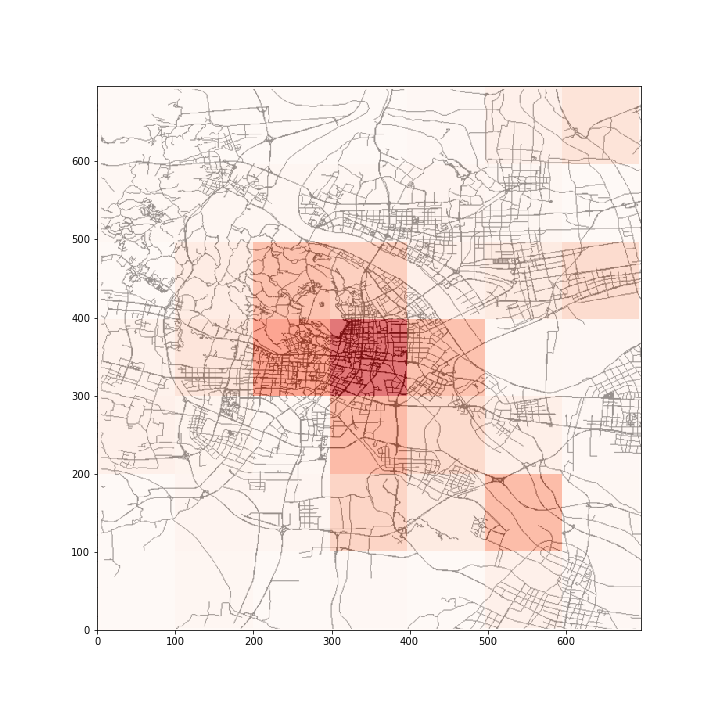}%
		\label{figure10d}}
	\\
	\subfloat[0-1 AM (P)]{\includegraphics[width=0.23\linewidth]{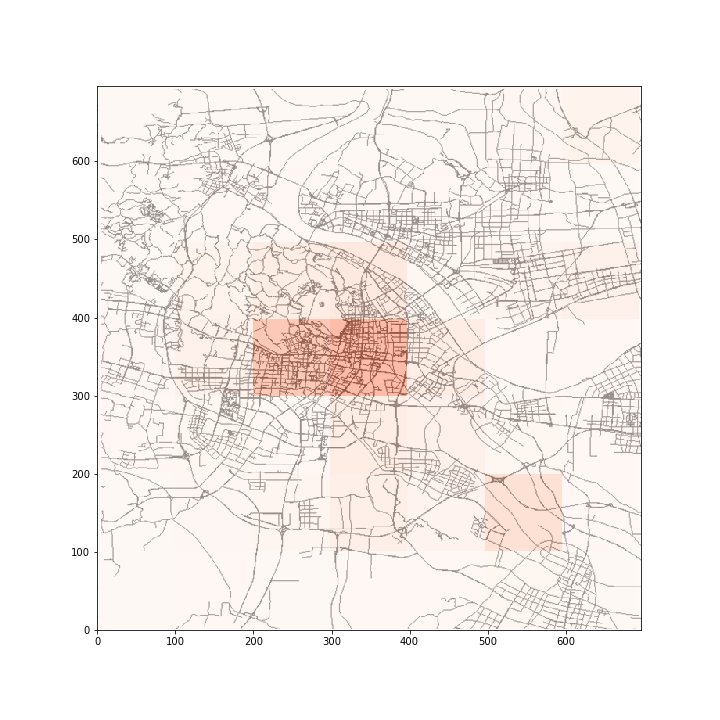}%
		\label{figure10e}}
	\hfil
	\subfloat[9-10 AM (P)]{\includegraphics[width=0.23\linewidth]{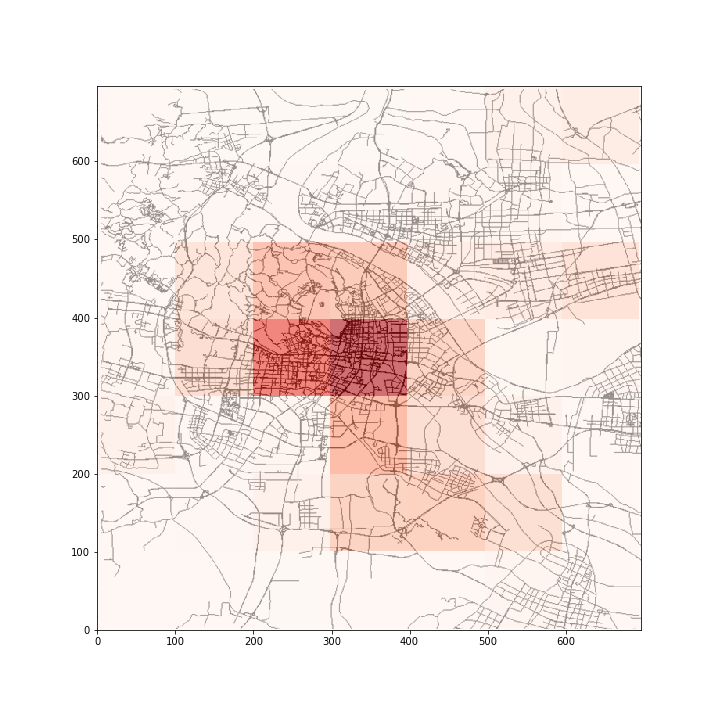}%
		\label{figure10f}}
	\hfil
	\subfloat[6-7 AM (P)]{\includegraphics[width=0.23\linewidth]{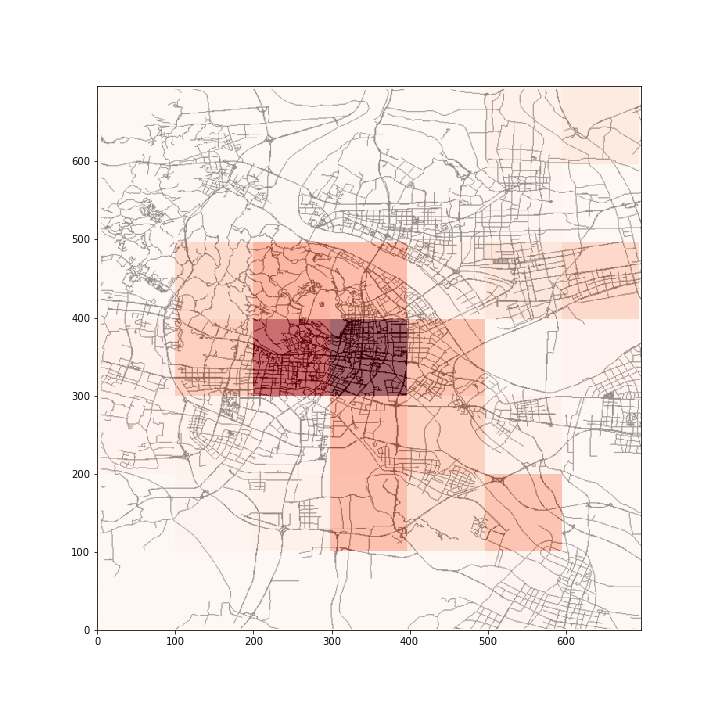}%
		\label{figure10g}}
	\hfil
	\subfloat[9-10 PM (P)]{\includegraphics[width=0.23\linewidth]{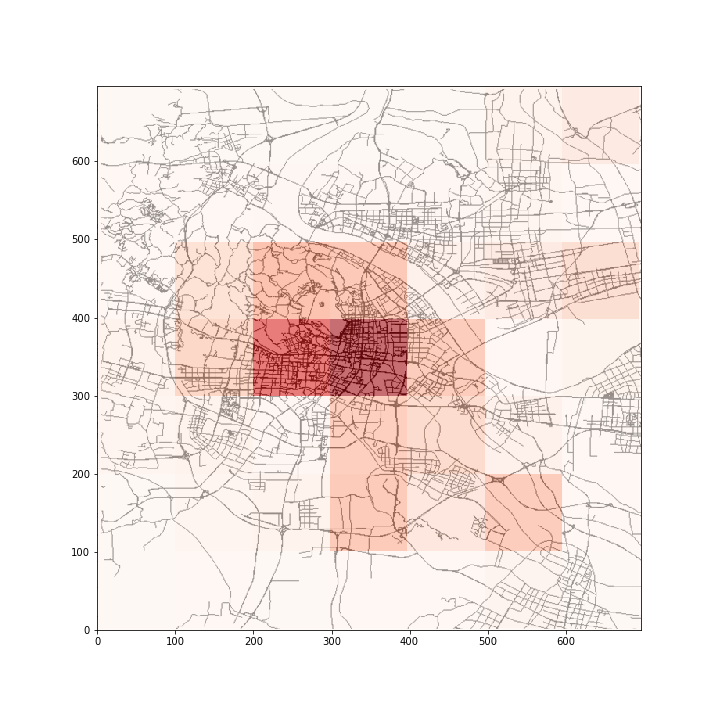}%
		\label{figure10h}}
	\caption{Comparison of the ground truth (G) and predicted passenger demand by FCL-Net (P).}
	\label{figure10}
\end{figure}

\section{Conclusions}

In this paper, we propose a DL approach, named the fusion convolutional LSTM (FCL-Net), for short-term passenger demand forecasting under an on-demand ride service platform. The proposed architecture is fused by multiple conv-LSTM layers, LSTM layers, and convolutional operators, and fed with a variety of explanatory variables including the historical passenger demand, travel time rate, time-of-day, day-of-week, and weather conditions. A tailored spatially aggregated random forest is employed to rank the importance of the explanatory variables. The ranking is then used for feature selection. We trained two FCL-Nets, one trained with full variables and another trained with selected variables. In addition, the conv-LSTM which only takes historical passenger demand as the explanatory variables is also established. These three models are compared with five benchmark algorithms including the HA, MA, ARIMA, ANN, and LSTM. The models are validated on the real-world data provided by DiDi Chuxing, the results of which show that the two FCL-Nets significantly outperform the benchmark algorithms, measured by RMSE, R-square, and MAE, indicating that the proposed approach performs better at capturing the spatio-temporal characteristics in short-term passenger demand forecasting. The FCL-Nets achieve approximately 50.9\% lower RMSE than the conv-LSTM (with only passenger demand as variables), implying that the consideration of the exogenous variables (such as the travel time rate, time-of-day, and weather conditions) is treasured. It is also interesting to find that the FCL-Net only suffers from 0.6\% loss in predictive performance (measured by RMSE) while the variable dimension of it is reduced by nearly 30\%, after feature selection. It indicates that appropriate feature selection helps reduce computaticcon complexity with little loss in the predictive accuracy.

This paper explores short-term passenger demand forecasting under the on-demand ride service platform via a novel spatio-temporal DL approach. Accurate real-time passenger demand forecasting can provide suggestions for the platform to rebalance the spatial distribution of cruising cars to meet passenger demand in each region, which will improve the car utilization rate and passengers' degree of satisfaction. However, to understand the complex interactions among the variables in the on-demand ride service market is far beyond predicting passenger demand. In the future, we expect to utilize more economic analyses and machine learning techniques to further explore the relationship of both the endogenous and exogenous variables under the on-demand ride service platform.

\section*{Acknowledgements}
This research is financially supported by Zhejiang Provincial Natural Science Foundation of China [LR17E080002], National Natural Science Foundation of China [51508505, 51338008], and Fundamental Research Funds for the Central Universities [2017QNA4025]. The authors are grateful to DiDi Chuxing (www.xiaojukeji.com) for providing us some sample data.

\section*{References}

\bibliography{articles}

\end{document}